\documentclass[journal]{IEEEtran}
\ifCLASSINFOpdf
  \usepackage[pdftex]{graphicx}
\else
\fi

\usepackage{amsmath}
\usepackage{subfigure}
\usepackage{lipsum}
\usepackage{siunitx}
\usepackage{hyperref} 
\usepackage{slashbox}
\usepackage{bm}
\begin{document}
%
\title{Delineating Bone Surfaces in B-Mode Images Constrained by Physics of Ultrasound Propagation}
%
%
%

\author{Firat~Ozdemir,
		Christine~Tanner,
        Orcun~Goksel
\thanks{F.\,Ozdemir, C.\,Tanner, and O.\,Goksel are with the Computer-Assisted Applications in Medicine Group, ETH Zurich, Switzerland. \mbox{(e-mail: ozdemirf@sdsc.ethz.ch, \{tanner,ogoksel\}@vision.ee.ethz.ch)}}}

\maketitle

\begin{abstract}

Bone surface delineation in ultrasound is of interest due to its potential in diagnosis, surgical planning, and post-operative follow-up in orthopedics, as well as the potential of using bones as anatomical landmarks in surgical navigation.
We herein propose a method to encode the physics of ultrasound propagation into a factor graph formulation for the purpose of bone surface delineation.
In this graph structure, unary node potentials encode the local likelihood for being a soft tissue or acoustic-shadow (behind bone surface) region, both learned through image descriptors.
Pair-wise edge potentials encode ultrasound propagation constraints of bone surfaces given their large acoustic-impedance difference.
We evaluate the proposed method in comparison with four earlier approaches, on in-vivo ultrasound images collected from dorsal and volar views of the forearm.
The proposed method achieves an average root-mean-square error and symmetric Hausdorff distance of 0.28\,mm and 1.78\,mm, respectively.
It detects 99.9\% of the annotated bone surfaces with a mean scanline error (distance to annotations) of 0.39\,mm.

\end{abstract}


%
\IEEEpeerreviewmaketitle

\section{Introduction}
%
%
%
%

National Ambulatory Medical Care Survey and American Academy of Orthopaedic Surgeons report that each year around 6.8 million patients seek medical attention for bone fractures in the US, among which almost 900,000 require hospitalization.
According to International Osteoporosis Foundation, approximately 1.6 million hip fractures occur worldwide and it is expected to increase 3 to 4 fold by 2050~\cite{Gullberg1997,Cooper1992}. 
Additionally, vast proportion of vertebral fractures are unrecognized; up to 45\% in the Americas and 29\% in Europe~\cite{JBMR5650200402}.
In Asia, underdiagnosis is even more significant due to the population mostly living in rural areas~\cite{mithal2009asian}, leading to limitations in access to hospitals and necessary diagnosis equipment. 
Due to the high vulnerability to orthopedic conditions in daily life, there is an increasing interest and focus on improving orthopedic imaging and other complementary technologies.

Bone surface localization and segmentation are beneficial for many orthopedic procedures including diagnosis, surgical planning, intra-surgical guidance~\cite{ciganovic2018registration}, and follow-up care. 
While accurate diagnosis is essential for best treatment, additional imaging contrast or visual anatomical aids would allow for better surgical planning and intra-operative guidance; both improving surgical outcomes and enabling less invasive surgical alternatives.

Although X-ray radiography or computed tomography (CT) can reveal orthopedic diagnostic information, they are also major sources of synthetic radiation~\cite{deGonzalez2004345}. 
Ionizing radiation should be particularly avoided for vulnerable patients, e.g.,\ children and pregnant women~\cite{BrennerEllistonHallEtAl2001}.
Furthermore, conventional clinical X-ray imaging shows insufficient contrast for certain muscoskeletal tissue conditions such as fibrosis~\cite{cady1983ultrasonic,Pillen2011MuscleUS}.
Ultrasound (US) may provide an additional or alternative imaging modality for diagnosis and treatment.
Albeit its low signal-to-noise ratio (SNR), US is a non-ionizing, real-time, low-cost, portable and hence widely accessible imaging option.

\begin{figure}
\centering
\includegraphics[width=0.5\linewidth]{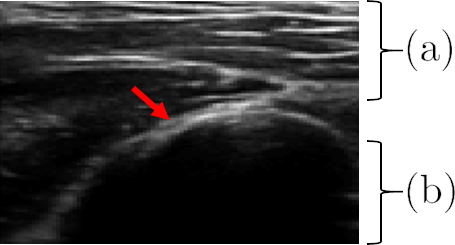}
\caption{Sample US image showing (a) layers of hyperechoic bands between soft tissue layers and the hyperechoic appearance (red arrow) of the bone surface, (b) the shadow below the bone surface.}
\label{fig:sampleUSimage}
\end{figure}

In US imaging, acoustic impedance difference between layers of tissues cause US signal to reflect, inducing the hyperechoic bands seen in B-mode images, see Fig.~\ref{fig:sampleUSimage}. 
Since the impedance difference between the bones and their surrounding tissues is relatively large, almost all incident acoustic power reflects, hence casting a shadow behind the bones, with any tissue below this point practically invisible as illustrated in Fig.~\ref{fig:sampleUSimage}. 
The thickness of the observed hyperechoic band at a bone surface depends on multiple factors including the imaging wavelength and the orientation of the US transducer relative to the bone surface~\cite{jain2004understanding}.
Although there are US radio-frequncy (RF) raw data based bone localization techniques, e.g.~\cite{wen2007Enhancement,hussain2014robust}, RF data access is not commonly available in commercial US systems, let alone its realtime streaming.
Therefore, most techniques in the literature utilize commonly available beamformed, demodulated, and dynamic-range adjusted \emph{brightness} (B)-mode images. 

Given the increased B-mode intensity values at bone surfaces, earlier studies in localizing bone surfaces use gradient or edge based approaches~\cite{daanen2004fully}.
\emph{Phase symmetry} (PS)~\cite{kovesi1999image} is the aggregation of US images filtered with different orientation and scale log-Gabor kernels; and PS was shown in~\cite{hacihaliloglu2009bone} to greatly emphasize the hyperechoic bands visible at bone surfaces.
Unfortunately, many hyperechoic bands between soft tissue layers show locally similar properties to bone surfaces, hence also yielding a high PS response, causing false positives. 
Despite its low specificity, PS has shown to provide satisfactory bone localization performance within manually selected regions of interest~\cite{hacihaliloglu2009bone}.
Later works expanded PS to 3D US~\cite{hacihaliloglu2015automatic}, automatized its parametrization~\cite{automaticHacihaliloglu2011}, and used phase tensors for registering statistical shape models to 3D US~\cite{hacihaliloglu2013statistical}.
With the aim of suppressing false detections, \emph{confidence in phase-symmetry} (CPS)~\cite{quader2014confidence} was proposed by combining PS with attenuation and shadowing maps extracted from US image, resulting in so-called \emph{confidence maps}~\cite{karamalis2012ultrasound}.
In~\cite{Baka2016}, a machine learning approach is suggested to estimate a likelihood map of bone surface, where the likelihood map is then used in the cost term for a dynamic programming delineation of the bone surface.
Later, in~\cite{baka2017random} additional features were extracted with a random forest classifier to segment bone surfaces. 
Further studies utilizing dynamic programming and combining phase tensors suggest additional improvements~\cite{hacihaliloglu2018localization}.
In~\cite{villa2018fcn}, a fully convolutional network (CNN) is trained on the confidence map~\cite{quader2014confidence}, PS, and the B-mode ultrasound images to delineate bone surfaces. 
Later, additional deep learning based methods fused with phase tensors were proposed to accurately segment bone surfaces~\cite{alsinan2019automatic,puyang2018simultaneous}.
In a recent study, state-of-the-art bone surface delineation accuracy is shown to be achieved with a standard CNN-based end-to-end learning approach~\cite{ciganovic2019deep}, hence substantially reducing the inference time.
It is noteworthy to emphasize on the influence of available datasets in works utilizing CNNs, where works with relatively smaller datasets ($300+$ images) report substantially inferior results when a CNN is trained using US images alone~\cite{alsinan2019automatic,puyang2018simultaneous}.

In this work, we model the physics of US propagation and interaction with bone surfaces for delineating such surfaces. 
We encode such information in a graph formulation, where we define the bone surface as the posterior interface between two graph labels for \textit{soft tissue} lying above and \textit{acoustic shadow} cast below the bone surface. 
Pairwise potentials are used with directed edges to encode US propagation, where \emph{vertical edges} encourage the label interface to align with the hyperechoic surfaces visible in the image, while enforcing an acoustically plausible order of labels.
In particular, when going downward, once the shadowing is initiated, then no tissue shall be visible anymore.
\emph{Horizontal edges} encourage spatial smoothness of the resulting labels and hence bone delineations.
Using a set of handcrafted features and annotated US images, we train binary classifiers for assigning unary potentials to the graph  nodes at each image pixel.
Solving this factor graph with directed edges,
the bone surface is delineated as the border between tissue and shadow labels. 
We compare this with our implementations of \cite{hacihaliloglu2013statistical} and~\cite{quader2014confidence}, as well as with our earlier work~\cite{ozdemir2016graphical}, which uses a separate additional label for the bone surface and \textit{jump edges} at all graph nodes to enforce a given thickness of bone surface appearance.
In contrast, we propose herein a novel bone surface definition that yields itself to a \emph{binary} graph labeling problem, validated through extensive evaluations and comparisons. 
We also investigate an extended set of features, such as local texture, curvature, and Haar-like features.

\section{Material}

\label{sec:experiments}

For evaluations, 37 US images from a diverse range of anatomical regions, including forearm (radius, ulna), shoulder (acromion, humerus tip), leg (fibula, tibia, malleolus), hip (iliac crest), jaw (mandible, rasmus) and fingers (phalanges) from one volunteer were collected, which we will refer to as \textit{diverseUS} dataset.
In addition, 415 US images from the dorsal and volar sides of the left forearm from another volunteer were collected, which will be referred to as \textit{forearmUS} dataset. 
Experimental data was acquired using a SonixTouch machine (Ultrasonix, Richmond, Canada) and an L14-5 linear transducer.
The former dataset (diverseUS) was bilinearly resampled for isotropic one pixel per wavelength resolution within the scope of our earlier work~\cite{ozdemir2016graphical}.
For both datasets, an imaging frequency of 6.66 or 10 Mhz was used depending on the body location. 
The collected B-mode images had a depth within $[30,50]$\,mm with a pixel resolution within $[0.064, 0.232]$\,mm.
Gold-standard (GS) bone surface delineations were annotated following the guidelines in~\cite{jain2004understanding}. DiverseUS dataset additionally includes soft tissue and acoustic shadow annotations.
Hence, diverseUS is used for all the training and hyperparameter optimization whereas forearmUS is used only for evaluations, without any further training or parameter tuning to also study generalizibility.

\begin{figure}
\centering
\includegraphics[width=\linewidth, height=1.65in]{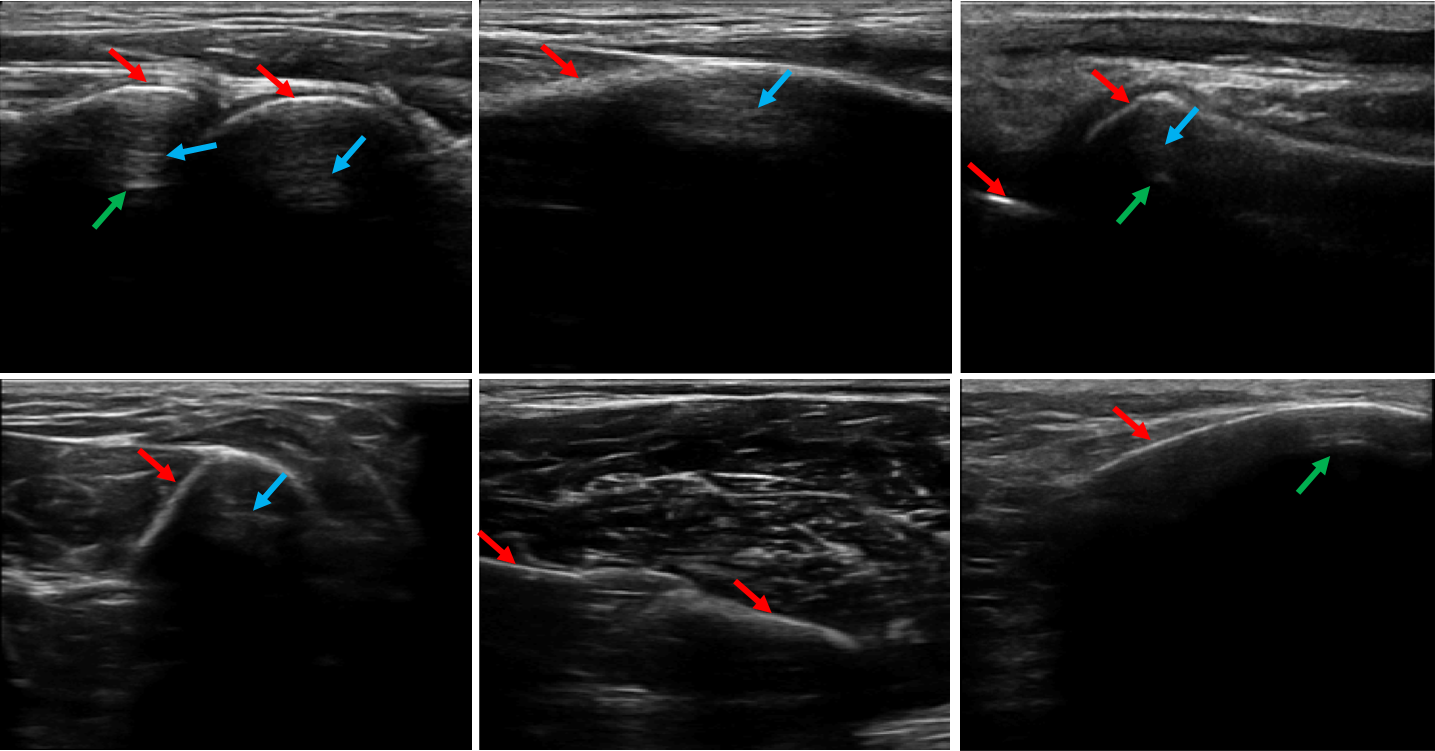}
\caption{Underneath the bone surfaces (red arrows), one can observe noise such as reverberations (blue arrows) or multiple reflections (green arrows) which may create ambiguity to infer shadowing and thus the bone surface above as the structure-of-interest.}
\label{fig:differenceLabels}
\end{figure}

\section{Methodology}
When localizing and segmenting bone surfaces on US, there are three different structures of interest; \textbf{B}one surface that is visible as a hyperechoic band perpendicular to the axis of US propagation, \textbf{S}hadow below a bone surface due to reflected and highly attenuated US signal, and remaining soft \textbf{T}issues that are not inside the bone nor are shadowed.
Below we first introduce the classifiers we use for determining the likelihood of the above classes per image region.
Then, we present the proposed factor graph structure and potentially applicable formulations to impose the physics of US propagation.

\subsection{Feature-based Classification of B-mode Image Pixels}
\label{sec:features}

Although both bone surfaces and soft tissue interfaces may appear similarly in US images, near complete reflection and posterior attenuation of US signal leads to a shadowed region underneath bone surfaces. 
Its identification in the presence of noise may become ambiguous as shown in Fig.~\ref{fig:differenceLabels}; for instance, shadow not appearing completely hypoechoic due to US artifacts such as multiple reflections and reverberations.
We resolve such ambiguities in a probabilistic, learning-based framework based on relevant US image features.
For this given binary classification problem, we use \emph{LogitBoost}~\cite{Friedman98additivelogistic}, which is a logistic additive method similar to AdaBoost, but requiring less computational power. 
As weak learners we use simple trees, leading to boosted \emph{decision forests} for the local classification task.

\begin{figure*}
\centering
\begin{tabular}{l}
    \includegraphics[width=\textwidth]{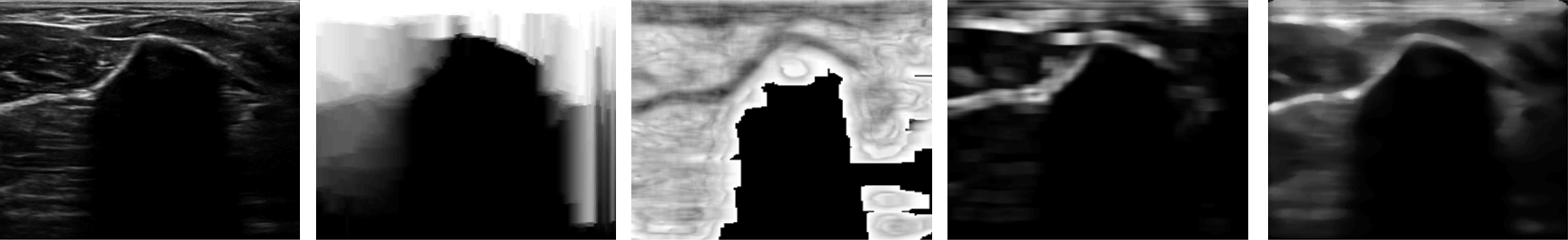}\\
    [-0.5ex]{%
    \hspace{1.25cm} image \hspace{1.9cm} confidence map \hspace{1.1cm} Rayleigh fit error \hspace{1.1cm} patch variance \hspace{1.25cm} patch median}\\
    \includegraphics[width=\textwidth]{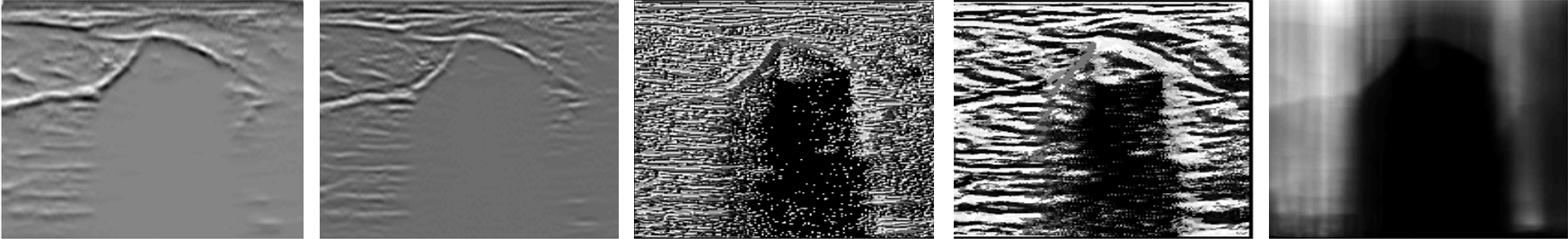}\\
    [-0.5ex]{%
    $1^{st}$ ord. CW local stat. \hspace{0.2cm} $2^{nd}$ ord. CW local stat. \hspace{1.35cm} LBP \hspace{2.2cm} extended LBP \hspace{0.8cm} CW cumulative mean}\\
    \includegraphics[width=\textwidth]{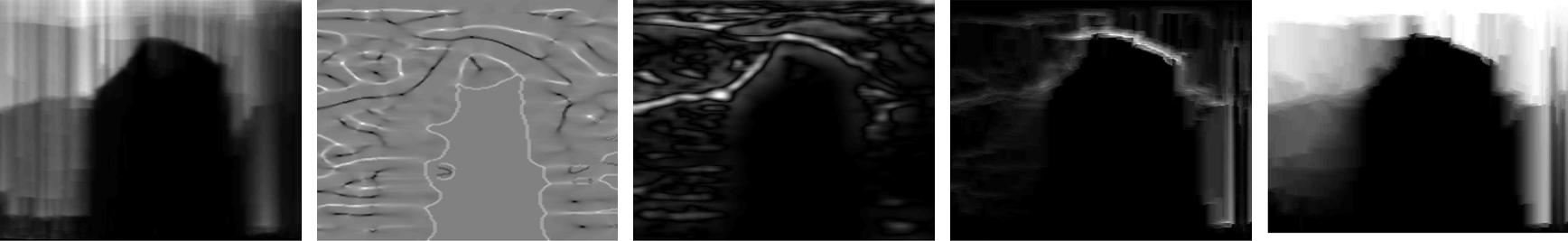}\\
    [-0.5ex]{%
    \hspace{0.25cm} CW cumulative std. \hspace{0.95cm} curvature map \hspace{1.35cm} Haar features \hspace{1.05cm} Attenuation feature \hspace{0.80cm} Shadowing feature%
    }
\end{tabular}
\caption{For an example US image, a representative subset of corresponding feature maps are shown.}
\label{fig:featureMaps}
\end{figure*}

Features are extracted at varying scales from 2D image patches as well as from 1D column vectors aligned with the propagation axis of US, as listed in Table~\ref{Table:FeatureList} and shown for a sample set in Fig.~\ref{fig:featureMaps}.
For this given binary classification problem, we use \emph{LogitBoost}~\cite{Friedman98additivelogistic}, which is a logistic additive method similar to AdaBoost, but requiring less computational power. 
As weak learners we use simple trees, leading to boosted \emph{decision forests} for the local classification task.
Features we utilize are described below.

\noindent
\textbf{Local Patch Statistics:}
Lower-order statistics such as mean and variance can separate \textbf{T} and \textbf{S} regions thanks to the speckle noise and various tissue-layer reflections present within the former but not the latter.
Additionally, entropy and higher-order statistics such as skewness and kurtosis may help with separating noise in \textbf{S} regions from \textbf{T}. 
Furthermore, we compute median, standard deviation, and energy (i.e.,\ sum of squared intensities) within local patches.
Eight features extracted at $3$ patch scales yield a total of $24$ features governing local patch statistics. 

\noindent
\textbf{Random-Walks based Statistics:}
Based on the random-walks framework~\cite{grady2006random}, confidence maps ($m_\mathrm{conf}$) from~\cite{karamalis2012ultrasound} aim to quantify the confidence in the information seen in a US image at pixel resolution. 
Assuming the top row of a US image to be in contact with the transducer, each pixel's virtual ``strength'' to reach the transducer is solved.
Confidence maps are computed through random walks on an undirected graph with higher edge costs for diagonal and horizontal connections, since only a few transducer elements are assumed to contribute to a given scanline of the US image with diagonal signal transmission being less likely.
Taking confidence maps as its basis, attenuation ($A_\mathrm{CPS}$) and shadowing ($S_\mathrm{CPS}$) metrics were proposed for enhancing bone surfaces in US~\cite{quader2014confidence}. 
Having confidence map $m_\mathrm{conf}$ for image $\mathbf{I}$, the attenuation metric at position $\mathbf{x}$ is given by
\begin{equation}
    A_{\mathrm{CPS}}(\mathbf{x}) = Z_\mathrm{A}^{-1} \sum_{i}\left(m_\mathrm{conf}(\mathbf{x}) - \min(m_\mathrm{conf}(w_i(\mathbf{x})))\right)
    \label{eqn:Attenuation}
\end{equation}
where $w_i(\mathbf{x})$ corresponds to the neighborhood patch {around $\mathbf{x}$ at scale $i$} and $\min(m_\mathrm{conf}(w_i(\mathbf{x}))))$ thus stands for the minimum of the confidence map value within $w_i(\mathbf{x})$. 
{$Z_\mathrm{A}$ is the normalizing factor} {to map $A_{\mathrm{CPS}}$ to [0, 1]}. 
Similarly, the shadowing metric is computed by
\begin{equation}
    S_\mathrm{CPS}(\mathbf{x}) = Z_\mathrm{S}^{-1} \sum_{i} \frac{ m_\mathrm{conf}(\mathbf{x})}{\min(m_\mathrm{conf}(w_i(\mathbf{x})))}
\label{eqn:Shadowing}
\end{equation}
where {$Z_\mathrm{S}$ normalizes $S_\mathrm{CPS}(\mathbf{x})$ to [0, 1].}
We include \{$m_\mathrm{conf}$, $A_\mathrm{CPS}$, $S_\mathrm{CPS}$\} along with $\log (S_\mathrm{CPS})$ in our feature space.

\noindent
\textbf{Column-wise (CW) Local and Cumulative Statistics:}
Since signal propagation is in the axial plane along the scanlines, we run different kernels along this axis to emphasize characteristics of different interfaces at varying scanlengths. 
Using Gaussian kernels at 3 different scales, we compute smoothed 1D gradients of up to 4 orders at a given image location along (vertical) US propagation axis, which we call CW local statistics.
These yield to $3$$\times$$4$=$12$ features from gradient convolutions.
While lower order characteristics look similar, combination of higher-order gradients can be important for differentiating the classes.

Although local characteristics are important, cumulatively observing all pixels along the scanline below a given point can be instrumental for determining the label of that pixel. 
For example, commonly hypoechoic regions such as blood vessels can be distinguished from shadow by observing any non-hypoechoic location below that region.
Therefore, we also apply cumulative scanline-long filters where all axial (vertical) information below a given pixel is summarized with $3$ column-wise features: sum, mean, and standard deviation of all pixels below the given pixel down to bottom of the image.
These 3 features are called cumulative statistics.

\noindent
\textbf{Local Binary Patterns (LBP):}
LBP~\cite{ojala1996comparative} feature is known to be a powerful texture descriptor~\cite{1717463}. 
Aside from low computational complexity, its invariance to monotonic intensity changes (e.g.,\ due to changing US imaging gain or time-gain-compensation settings) makes LBP attractive for US texture discrimination.
LBP descriptors are typically represented in 8 bits, where the intensity of each pixel is thresholded with its 8-connected immediate neighbors in a consistent order to generate a binary vector.
In~\cite{froba2004face}, Modified Census Transform (MCT) was proposed for a robust threshold value as the mean of a $3$$\times$$3$ neighborhood, adding invariance to illumination change and noise. 

In addition to the $3$$\times$$3$ pixel region used in LBP and MCT, we propose to also include every second pixel in each $5$$\times$$5$ neighborhood like in a checker-board pattern.
Then, for an \emph{extended LBP} feature, the differences of pixel pairs that are symmetric with respect to the center pixel are thresholded by 0. 
For an \emph{extended MCT}, the differences of these opposing pairs are thresholded with the difference between the center pixel intensity and the mean intensity of the 25 pixels in the given neighborhood.
These features remain invariant to slow intensity changes while being sensitive to spikes at multiple orientations and widths.
These small variations could be descriptive when discriminating {bright reflections at bone surfaces from soft tissue and shadow regions.}
These 4 features above are extracted at a single scale.

\noindent
\textbf{Speckle Characterization:}
In B-mode images, even in the absence of specular reflections, coherent noise from scattering cause the typical US speckle texture.
Fully-developed speckle intensities can be characterized by Rayleigh, Nakagami, or distributions of similar nature.
Such characterization may help to differentiate tissue speckle ``noise'' (texture) from other potential noise source, e.g.,\ in shadowed regions \textbf{S}, the latter likely containing different distributions due to different nature of origin, e.g.,\ electric noise.
Accordingly, we use the fitness of a patch histogram to Rayleigh distribution as a feature as follows:
\begin{equation}
\mathrm{FIT_R}(\mathbf{x}) = ||\ \mathrm{hist}(w(\mathbf{x})) - \mathrm{pdf_{R}}(w(\mathbf{x})) \ || 
\end{equation}
where the first term is the normalized histogram of the pixel intensities within patch $w$ centered at $\mathbf{x}$, and the second term $\mathrm{pdf_{R}}$ is the maximum-likelihood fitted distribution to this histogram.
We implemented $\mathrm{pdf_{R}}$ in Matlab as
\begin{equation}
\mathrm{pdf_{R}}(w(\mathbf{x})) = \mathrm{raylpdf}(w(\mathbf{x}) | \mathrm{raylfit}(w(\mathbf{x})))
\end{equation}
where $\mathrm{raylfit(\cdot)}$ is the maximum likelihood estimation of the Rayleigh scale parameter for the given data~\cite{johnson2005univariate} and $\mathrm{raylpdf(\cdot|\cdot)}$ is the Rayleigh probability density function~\cite{evans2000statistical} for the given parameter.
Due to high computational demand, we used a single patch size, leading to a single feature.

\noindent
\textbf{Local Texture Response:}
Spatio-temporal {information} from an image can be extracted using Gabor filters tuned at different frequencies and spatial orientations of interest.
For filter frequency $f$ and orientation $\theta$, a set of Gabor filters
\begin{equation}
G(\mathbf{x}, \bm{\sigma}, f, \theta) = g(\mathbf{x}, \bm{\sigma}) \cos(2 \pi f (x_{1} \cos(\theta) + x_{2} \sin(\theta)))
\end{equation}
can be defined, where $g(\mathbf{x}, \bm{\sigma})$ is the 2D zero-mean Gaussian functions characterized by different variances $\bm{\sigma}$, and $x_i$ are the components of $\mathbf{x}$.
A Gabor filtered image, called the \emph{local texture map}, is included among our features.

\noindent
\textbf{Local Curvature Response:}
Image curvatures may be useful for discriminating soft tissue layers and  bone surfaces.
We include image curvature in 2D as a feature as defined in~\cite{schnabel1995active} as
\begin{equation}
K(\mathbf{x}) = - \nabla \cdot  \frac{ \nabla \mathbf{I}(\mathbf{x}) }{||\nabla \mathbf{I}(\mathbf{x})||}  
\end{equation}
where $\nabla \cdot$ and $\nabla$ are divergence and gradient operators, respectively. 

\noindent
\textbf{Haar-Like Features:}
Haar-like features~\cite{viola2001rapid} are widely used in fast object detection tasks, thanks to their fast computation by exploiting integral image properties.
We extract Haar-like features at 5 different scales in the form of \emph{center-surround features}~\cite{1038171}, where a smaller square is subtracted from a larger co-centric square.  

\begin{table}[!b]
\caption{List of 56 features extracted at different kernel space-scales for US transmit wavelength ($\lambda$) and pixels (px).}
\label{Table:FeatureList}
\centering
\scriptsize
\begin{tabular}{@{\hskip 0mm}l@{\hskip 1mm}l@{\hskip 0mm}c@{\hskip 0.5mm}r@{\hskip 0mm}}
{\bf Feature Types} & {\bf Feature Group Names} & {\bf Scale} & {\bf \#}\\ 
\hline
Intensity & Pixel intensity & 1px & 1\\
\hline
Local patch statistics & \begin{tabular}[c]{@{}l@{}}Mean, median, variance, standard deviation\\skewness, kurtosis, entropy, {energy}\end{tabular} & 3,6,12$\lambda$  & 24\\
\hline
Random Walks & \begin{tabular}[c]{@{}l@{}}Confidence Map~\cite{karamalis2012ultrasound}, \\ Shadowing, $\log$Shadowing, Attenuation~\cite{quader2014confidence} \end{tabular} & \begin{tabular}[c]{@{}c@{}} 1px \\ 2,5,11$\lambda$ \end{tabular} & 4\\
\hline
Column-wise (CW) local & {CW local statistics (order $\in$$\{0,1,2,3\})$} 
& 5,11,31$\lambda$ & 12\\
\& cumulative statistics & {Sum}, mean, standard deviation & - & 3 \\
\hline
Local Binary Patterns& \begin{tabular}[c]{@{}l@{}}{(normal~\&}~{extended)} Local Binary Patterns\\ \cite{ojala1996comparative} and \\Modified Census Transform~\cite{froba2004face}\end{tabular} & - & 4\\
\hline
Speckle characteristics & {Rayleigh fit error} & 12$\lambda$ & 1\\
\hline
Local texture & {Gabor filter response} {($\theta$=$\pi/2$ $f$=$16$Hz)} & $\bm{\sigma}$=$[2,4]$mm  & {1} \\ 
\hline
Local curvature & {Curvature response} & {-} & {1} \\
\hline
Haar-like features & {Center-surround Haar features} & {2,5,9,15,25,30px} & {5} \\
\end{tabular}
\end{table}

\subsection{Graph Formalization of US Interaction with Bone Surface}
\label{sec:pairwiseConnections}

Although the learned classifiers above may already suppress some false-positive detections and false-negative delineation gaps, these utilize potentially ambiguous and sometimes contradictory local image descriptions.
Indeed, incorporating global image context can help in resolving local ambiguities, and hence is essential for resolving false bone surface detections.
For instance, a tissue \textbf{T} appearance inside a shadow \textbf{S} region can only be explained and resolved by either concluding the appearance being an artifact in the shadow (hence correcting the apperance to label \textbf{S}) or realizing that the falsely-presumed bone surface supposedly casting that shadow is potentially a tissue interface instead (hence correcting the region lying above the actual shadow to label \textbf{T}).
CPS is an approach to combine PS with global US properties through linear weighting.
Although this may emphasize bone surfaces, the resulting segmentation performance was found in~\cite{ozdemir2016graphical} to be inferior to PS, due to CPS responding only when there already is a PS response. 
Alternatively, global contextual constraints can be incorporated naturally and more accurately through a graph formalization of this segmentation problem.
Below we describe two alternative approaches; with and without an explicit bone surface label, where the nature and advantages are comparatively described and evaluated in the following sections.

\begin{figure}
\centering
\includegraphics[width=0.92\linewidth]{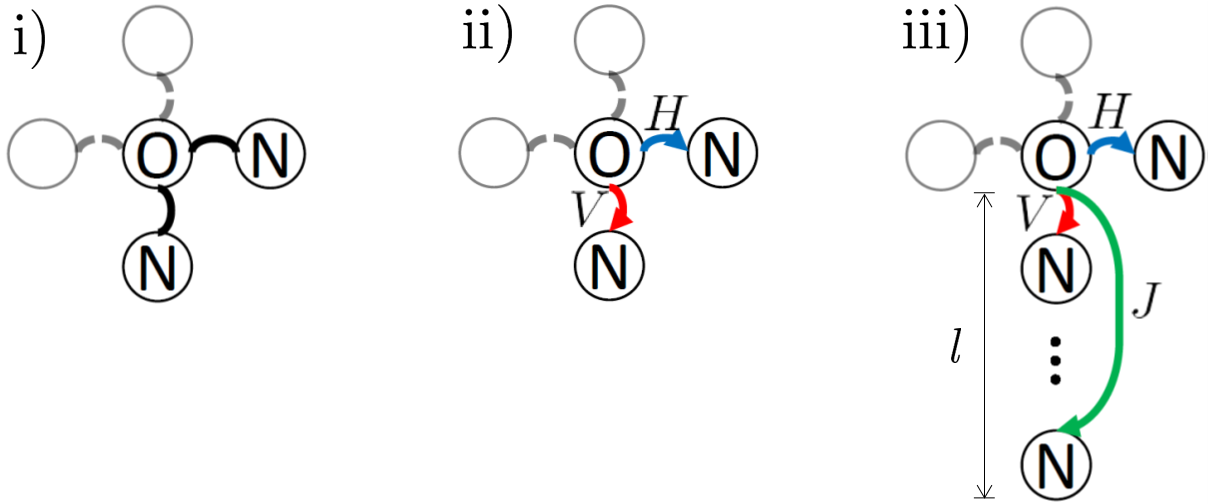}
\caption{
Pairwise edge connection representations for (i) the Potts model, and our proposed (ii) directed lateral and scanline edges as well as (iii) bone factor graph (BFG). 
}
\label{fig:PairwiseConnections}
\end{figure}

\noindent{\bf Graphical Model and Pairwise Edge Potentials. }
We represent the US image as a Markov random field (MRF) with the following energy
\begin{equation}
    \sum_{i} \Psi(i) \ + \ \mu \sum_i \sum_{j \in \mathrm{N}_{i}} \Psi(i, j)
\end{equation}
where $\Psi(\cdot)$ and $\Psi(\cdot, \cdot)$ are the unary nodal and pairwise edge cost functions and $\mathrm{N}_i$ is the neighbourhood of node $i$.
A 4-connected edge configuration is used for the graph representation and connectivity of the pixels in the image as shown in Fig.~\ref{fig:PairwiseConnections}(i).

For spatial regularization of labels, a trivial option is a Potts-like pairwise model, where any edge $e(i,j)$ connecting nodes $i$ and $j$ will be penalized higher when the labels differ, i.e.\ $l(i)$$\neq$$l(j)$.
This is seen in Table~\ref{tbl:pairwiseCost_BFG}(a) when $k_1$$<$$1$ and $k_2$$<$$1$.
This table assumes directed edges. 
For MRFs with undirected edges, one can consider only the lower triangle of $\Psi^\mathrm{H}(i,j)$.

\begin{table*}
\setlength\tabcolsep{0.5pt}
\caption{Pairwise cost definitions for (a) horizontal \textbf{H}, (b) vertical \textbf{V}, and (c) jump \textbf{J} edges. 
$\infty_{1}$ discourages undesired label neighbourhood, which would disobey the physcially-expected \textbf{T}$\rightarrow$\textbf{B}$\rightarrow$\textbf{S} label order. 
$\infty_{2}$ ensures a bone surface before shadow. 
$\infty_{3}$ and $\infty_{4}$ ensure a limited thickness for the bone surface. 
$k$ parameters control the spatial regularization.}
    \label{tbl:pairwiseCost_BFG}
    \centering
    \footnotesize
    \begin{tabular}{@{\extracolsep{\fill}}c c c c}
        (a) & (b) & (c) &\\
    \begin{tabular}{c@{\ }|@{\ }c@{\ }|@{\,}c@{\,}|@{\,}cccc}
                        $\Psi^\mathrm{H}(i,j)$ & {\textbf{T}}$(j)$ & {\textbf{B}}$(j)$ & {\textbf{S}}$(j)$ \\
                        \cline{1-4}
                        {\textbf{T}}$(i)$ & $k_{1}$ & 1 & 1 & \\
                        \cline{1-4}
                        {\textbf{B}}$(i)$ & 1 & $k_{2} $ & 1 & \\
                        \cline{1-4}
                        {\textbf{S}}$(i)$ & 1 & 1 & $k_{1}$ &
    \end{tabular}
    \hspace{0.8em}
    &
    \begin{tabular}{c@{\ }|@{\ }c@{\ }|@{\,}c@{\,}|@{\,}cl}
                        $\Psi^\mathrm{V}(i,j)$ & {\textbf{T}}$(j)$ & {\textbf{B}}$(j)$ & {\textbf{S}}$(j)$ &  \\
                         \cline{1-4}
                        {\textbf{T}}$(i)$ & $k_{2}  $  & $k_{3}$ & $\infty_{2}$ &  \\
                         \cline{1-4}
                        {\textbf{B}}$(i)$ & $\infty_{1}$ & $k_{2} $ & $k_{3}$ & \\
                         \cline{1-4}
                        {\textbf{S}}$(i)$ & $\infty_{1}$ & $\infty_{1}$ & $k_{2}$ &
    \end{tabular}
    \hspace{0.8em}
    &
    \begin{tabular}{c@{\ }|@{\ }c@{\ }|@{\,}c@{\,}|@{\,}c}
                        $\Psi^\mathrm{J}(i,j)$ & {\textbf{T}}$(j)$ & {\textbf{B}}$(j)$ & {\textbf{S}}$(j)$\\
                        \cline{1-4}
                        {\textbf{T}}$(i)$ & 0 & 0 & $\infty_{3}$\\
                        \cline{1-4}
                        {\textbf{B}}$(i)$ & $\infty_{1}$ & $\infty_{4}$ & 0 \\
                        \cline{1-4}
                        {\textbf{S}}$(i)$ & $\infty_{1}$ & $\infty_{1}$ & 0 
    \end{tabular}
    &
    \hspace{0.8em}
    \begin{tabular}{l}
    \scriptsize
                        
                        $k_1 \in [0.1,1]$\\
                        $k_2 \in [0.1,1]$\\
                        $k_3 \in [0.1,10^3]$\\
                        $\infty_{i}$=$10^4$
  \end{tabular}
    \end{tabular}
\end{table*}

\subsubsection{US Propagation Edges}

In order to enforce the strict propagation prior of US, lateral edges, also referred to as horizontal (\textbf{H}), and edges along the US scanline, similarly referred to as vertical (\textbf{V}), are herein proposed to have different pairwise energy definitions.
For no bone in the scanline, all nodes should be tissue \textbf{T}.
For a top-to-bottom traversal of the US image in depth, if and once a bone surface is encountered, the labels in this said scanline should strictly follow the order: \textbf{T}, \textbf{B}, and \textbf{S}, since after the bone surface only shadow can be visible.
Such an order can be enforced in a graphical model, only using directed edges; i.e.,\ allowing $e(i,j) \neq e(j,i)$.  
For defining such directed edges, we use factor graphs~\cite{kschischang2001factor}, with which we encode the above-mentioned scanline order as a propagation prior using direction-dependent weights between vertically-neighbouring nodes.

With the motivation above, a pairwise cost scheme for vertical neighbours shown in Table~\ref{tbl:pairwiseCost_BFG}(b) was proposed in~\cite{ozdemir2016graphical}, summarized below for completeness.
Large penalties (represented with $\infty_1$) prohibit undesired vertical neighbourhoods; i.e.\ preventing \textbf{B}$\rightarrow$\textbf{T}, \textbf{S}$\rightarrow$\textbf{B}, and \textbf{S}$\rightarrow$\textbf{T} assuming directed edges pointing the US propagation direction (downward).
$\infty_2$ ensures a layer of bone surface \textbf{B} to exist at any \textbf{T}$\rightarrow$\textbf{S} transition.
Parameters $k_2$ and $k_3$ allow to weight encountering a bone interface with respect to the continuation of the same label as shown in Table~\ref{tbl:pairwiseCost_BFG}(b).
For the horizontal edges, no meaningful physics priors can be foreseen, but the anatomical connectivity can be utilized as a geometric prior for spatial regularization with Potts potentials where homogeneity, i.e.,\ $e(i,j)$ can be penalized lower when $i=j$ than otherwise, as seen in Table~\ref{tbl:pairwiseCost_BFG}(a). 
Here we defined a separate penalty $k_1$ for \textbf{S} and \textbf{T} pixels compared to $k_2$ for \textbf{B}, since a large class imbalance is expected.
Note that in a typical image there would be large regions of shadow \textbf{S} and tissue \textbf{T}, in comparison to bone surface \textbf{B} pixels. 
Therefore, we use a larger horizontal weight $k_2$ for bone surface, which requires stronger agreement among unary potentials for continuity in the lateral direction.

\subsubsection{Jump Edge}

Despite the above constraints, graph solutions with a surface stretching over several centimeters thickness (e.g.,\ with two or more layers of tissue reflections being combined as a thick bone surface) are still valid solutions. 
In contrast, in B-mode images the bone surfaces appear as hyperechoic bands of finite thickness, due to the limited (pulse-length related) axial resolution of ultrasound among other factors~\cite{jain2004understanding}.
Accordingly, we enforce bone surface \textbf{B} to vertically be a finite-width layer, the thickness $l$ of which can be defined as a function of US frequency $\lambda$.
In order to enforce such given thickness, we use additional \emph{jump} (\textbf{J}) edges that connect each node to another one precisely $l$ nodes below, as shown in Fig.~\ref{fig:PairwiseConnections}(iii) and with the costs given in Table~\ref{tbl:pairwiseCost_BFG}(c).
A large penalty $\infty_3$ (similarly to $\infty_2$) prohibits any direct transition from \textbf{T} to \textbf{S} within $l$ nodes, thus requiring a label \textbf{B} at an horizon of $l$ -- effectively putting a minimum vertical thickness constraint of $l+1$ on the posterior of \textbf{B}.
Additionally, $\infty_4$ prevents the two ends of a \emph{jump edge} being both \textbf{B}, effectively constraining the maximum vertical thickness of any posterior \textbf{B} band to $l+1$.
These two above constraints combined ensure all detected bone surfaces \textbf{B} to be exactly $l$ pixels thick vertically.
The cost term $\infty_1$ in Table~\ref{tbl:pairwiseCost_BFG}(c) acts similarly and concurrently to $\Psi^\mathrm{V}$ to ensure the logical order of \textbf{T}$\rightarrow$\textbf{B}$\rightarrow$\textbf{S}.

The factor-graph model combining \textbf{J}, \textbf{V}, and \textbf{H} edge definitions as shown in Fig.~\ref{fig:PairwiseConnections}(iii) for bone surface segmentation in US images was called \emph{bone factor graph} (BFG) in~\cite{ozdemir2016graphical}. 
This, however, has several shortcomings in design, paramterization, and computation:
First, the bone thickness parameter $l$ above is hard-coded into the graph formulation and thus it needs to be empirically defined a priori, which is not a trivial task.
Moreover, this surface thickness $l$ would in practice not be invariant, but would depend on depth from surface, focusing depth, surface inclination, aberrations, image post-processing operations, etc. 
Furthermore, the given jump edges above increase the total number of edges by $\approx 50\%$. 
Indeed, given that the actual bone surface (not the US appearance) is a layer of infinitesimal thickness, any need for a separate label \textbf{B} to that end is actually questionable. 
To address these points we herein propose an alternative simplified graph structure, where the bone surface is encoded in the edges rather than a separate label, as described below.

\subsubsection{Bone-Responsive Edges}
 
\begin{figure}
\centering
\includegraphics[trim = 0mm 0mm 0mm 0mm, clip, width=\linewidth]{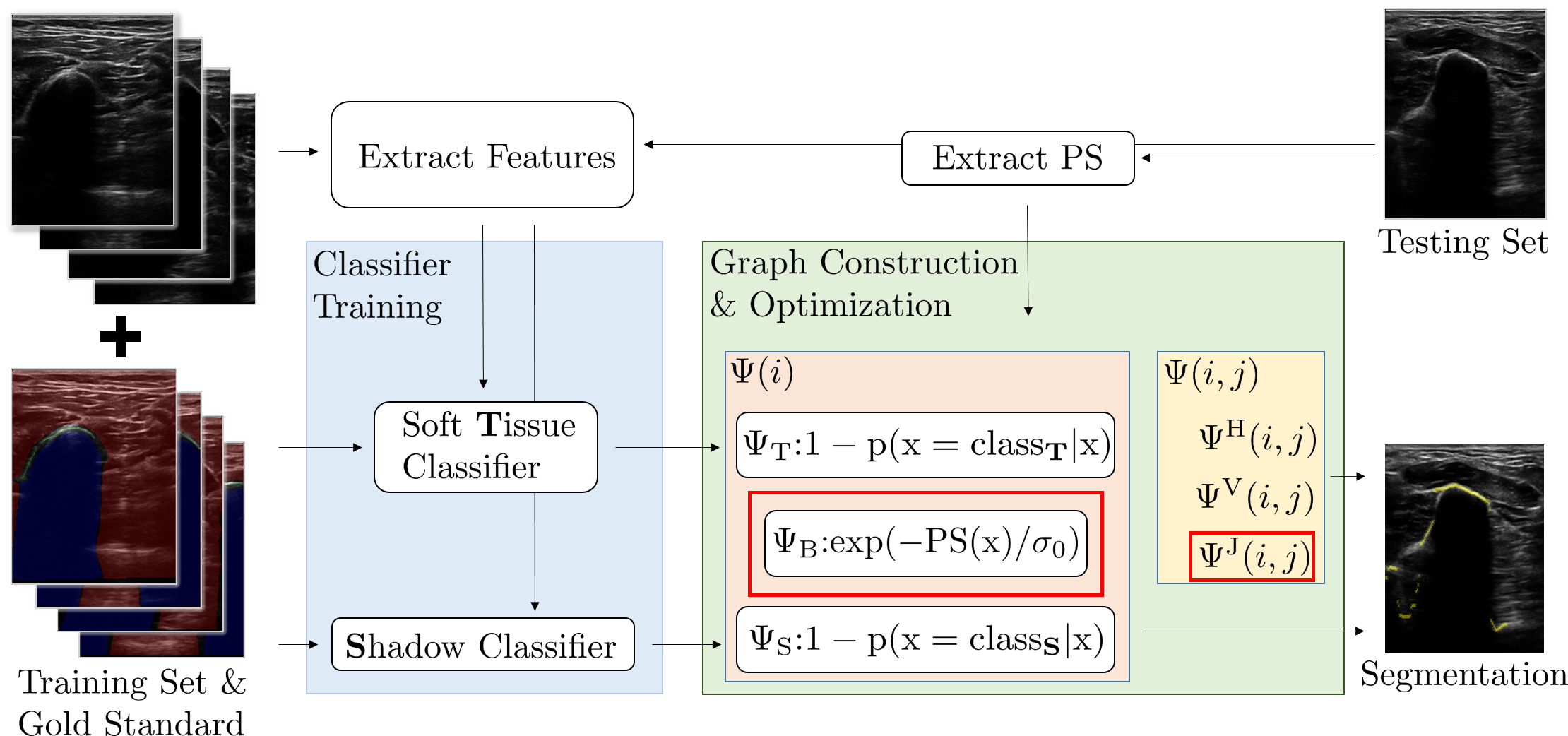}
\caption{Overview of the presented framework CBG, where the components within the red contour are valid only for the proposed BFG configuration.}
\label{fig:frameworkSchematic}
\end{figure}

We modify the above model by removing the jump edges as well as discarding the label \textbf{B}, and instead encoding bone surface likelihood on vertical \textbf{T}$\rightarrow$\textbf{S} edges.
This yields a simplified directed graph as seen in Fig.~\ref{fig:PairwiseConnections}(ii).
In this work, we use the pixel-wise PS response to define the bone-responsive edge potentials locally; i.e.,\ the vertical edge potential is redefined as $\Psi^\mathrm{V}(i,j)$$=$$k_3f_\mathrm{PS}(\mathbf{x}_i) $  for $l(i)$$=$\textbf{T} and $l(j)$$=$$\textbf{S}$, where $k_3$ is a weighting constant as shown in Table~\ref{tbl:pairwiseCosts_sBFG} and
\begin{equation}
f_\mathrm{PS}(\mathbf{x}_i) = e^{\frac{\mathrm{PS}(\mathbf{x_i})}{-Z_\mathrm{PS} \sigma_0}}
\end{equation}
where $\mathrm{PS}(\mathbf{x}_i)$ is the phase-symmetry response at the image position $\mathbf{x}_i$ of node $i$, $Z_\mathrm{PS}$ is the normalization factor to scale PS values across a given image to the range $[0,1]$, and $\sigma_0$ regulates an exponential decay. 
For horizontal edges we use the earlier Potts-like spatial smoothing model, as shown in Table~\ref{tbl:pairwiseCosts_sBFG}(a).
Since the vertical edges are conditioned on a spatially varying function (PS, in this scenario) as in conditional random fields, we refer to our proposed model above as \emph{conditional bone graph} (CBG).
An overview of CBG framework is shown in Fig.~\ref{fig:frameworkSchematic}.

\subsection{Implementation}
PS is computed as the sum over different {orientations $r$$=$$1,...,N_r$} and scales {$m$$=$$1,...,N_m$} as follows~\cite{hacihaliloglu2009bone} 
\begin{equation}
\mathrm{PS}(\mathbf{x}) = \frac{\sum_r \sum_m \lfloor ( |e_{r,m}(\mathbf{x})| - |o_{r,m}(\mathbf{x})| ) - t_r \rfloor}{\sum_r \sum_m \sqrt{e_{r,m}^2(\mathbf{x}) + o_{r,m}^2(\mathbf{x})} + \epsilon}
\end{equation}
where $t_r$ is an orientation dependent noise threshold, $\lfloor a \rfloor$$=$$\max(a,0)$, and $e_{r,m}$ and $o_{r,m}$ are, respectively, the real and imaginary components of image $\mathbf{I}$ filtered (convolved) with the 2D Log-Gabor filter $G_{r,m}(\omega, \phi)$. In other words,
\begin{equation}
e_{r,m}+jo_{r,m} = \mathbf{I} * (F^{-1}(G_{r,m}(\omega, \phi))) \,,
\end{equation}
{where $j$$=$$\sqrt{-1}$ is the imaginary unit,} $F^{-1}$ represents the inverse Fourier transform operator, and $G_{r,m}(\omega, \phi)$ can be customized in spectral domain as
\begin{equation}
G_{r,m}(\omega, \phi) = \exp\left(-\frac{(\log ( \omega / \omega_0))^2}{2(\log (\kappa_m / \omega_0))^2} {-} \frac{(\phi - \phi_r)^2}{2 {\sigma_\phi^2}} \right)
\end{equation}
where parameters $\phi_r$, $\omega_0$, {$\kappa_m$, and $\sigma_\phi$} allow to define the filter orientation, center frequency, scaling factor, and angular bandwidth of the band-pass filter, respectively.

\begin{table}
\caption{Pairwise cost definitions for horizontal \textbf{H} and vertical \textbf{V} edges for the proposed conditional bone graph (CBG).
$\infty_{1}$ discourages undesired label neighbourhood, which would disobey the physcially-expected \textbf{T}$\rightarrow$\textbf{S} label order. 
$k$ parameters control the spatial regularization.
$f_\mathrm{PS}(\cdot)$ embeds the local bone surface likelihood for regulating \textbf{T}$\rightarrow$\textbf{S} transitions.}
    \label{tbl:pairwiseCosts_sBFG}
    \centering
    \begin{tabular}{c@{\hskip 1cm}c}

	\begin{tabular}{c@{\ }|@{\ }c@{\ }|@{\,}c}
                        $\Psi_{\scriptscriptstyle \mathrm{CBG}}^\mathrm{H}(i,j)$ & {\textbf{T}}$(j)$ & {\textbf{S}}$(j)$ \\
                        \hline
                        {\textbf{T}}$(i)$ & $k_1$ &  1 \\
                        \hline
                        {\textbf{S}}$(i)$ & 1  & $k_1$
	\end{tabular}
	& 
	\begin{tabular}{c@{\ }|@{\ }c@{\ }|@{\,}c}
                        $\Psi_{\scriptscriptstyle \mathrm{CBG}}^\mathrm{V}(i,j)$ & {\textbf{T}}$(j)$ & {\textbf{S}}$(j)$ \\
                        \hline
                        {\textbf{T}}$(i)$ & $k_2$ & $k_3 f_\mathrm{PS}(\mathbf{x}_i)$ \\
                        \hline
                        {\textbf{S}}$(i)$ & $\infty_{1}$  & $k_2$
	\end{tabular}
    \end{tabular}
\end{table}

Given these potential definitions, we seek the maximum a posteriori (MAP) solution for a given graphical model.
Note that an exact inference may not always be possible, since submodularity cannot be guaranteed with conditional edge potentials and the directed edges prevent the use of many graph solvers.
Accordingly, all graphical models given above were implemented in OpenGM and solved using Sequential Tree-Reweighted Message Passing (TRW-S), which uses a dual ascent algorithm and allows for an approximate inference regardless of problem definition.

\section{Experimental Setup and Evaluation}
\label{sec:performanceTuning}

For evaluations and comparisons, we optimized the proposed algorithms in multiple aspects, including hyperparameter and feature selection (details are provided in the Appendix).

\subsection{Performance Metrics}
\label{sec:performanceMeasures}

As gold-standard (GS) bone surface for evaluations, only the bone surfaces visible with sufficient confidence were annotated.
Hence, non-annotated GS image columns (US scanlines) indicate no visible bone surface, either due to the actual in-existence of a bone or due to a potential ambiguity in its existence or confident localization.
Based on such \emph{under-segmentation} as GS, we compute common metrics such as root-mean-square error (RMSE), mean Euclidean distance (MED) and one-way Hausdorff distance (oHD) only from the GS with guaranteed bones to the automatic delineation with potential errors.
Furthermore, in order to compute symmetric Hausdorff distance (sHD), we evaluate delineations only for US scanlines where GS annotations exist. 
Provided that Hausdorff distance can be too prohibitive, we also report $95$-percentiles for oHD and sHD; denoted as oHD$^*$, and sHD$^*$, respectively.

Although RMSE, MED, oHD, and sHD measure different aspects of segmentation performance, we found them to be incomplete for assessing bone surface delineations with respect to GS annotations.
With RMSE and oHD, major errors at some scanlines can be compensated with more accurate segmentations found at neighboring scanlines; and sHD may be unfair by introducing a very large penalty, e.g.,\ for a single falsely-detected pixel at a far distance.
To assess the segmentation performance for each scanline separately, we therefore utilize two additional metrics: 
For each scanline where both manually (GS) and automatically segmented bone surface exists, the average distance error to all predictions along the scanline is measured, and their mean over the image is reported as \emph{mean scanline error} (MSE).
To quantify false detections, i.e. \emph{false negative rate}, we report the ratio of the number of scanlines with only GS but no delineation output to the total number of scanlines with GS annotations, called herein the \emph{miss percentage} (MP).

\subsection{Parameter Optimization}
\label{sec:parameterOptimization}

After a set of initial empirical tests and in view of earlier works such as~\cite{hacihaliloglu2009bone, automaticHacihaliloglu2011}, we fixed the parameters regarding Log-Gabor filter scale ($N_m$$=$$1$), ratio $\kappa / \omega_0$$=$$ 0.25$, and ensemble tree size ($t_\mathrm{size}$$=$$50$).
For the remaining 8 parameters of CBG we ran a grid-based parameter optimization with respect to  bone surface delineation. 
These 8 parameters are the number of orientations in PS computation ($N_r$), $f_\mathrm{PS}$ decay parameter ($\sigma_0$), {$f_\mathrm{PS}$ angular frequency ($\omega_0$) which we reported in spatial units as $\lambda_\mathrm{PS}$$=$$2 \pi / \omega_0$,} weighting coefficient of pairwise potentials ($\mu$), and pairwise edge cost parameters ($k_1, k_2, k_3$), and half the angle of coverage ($\angle_\mathrm{PS}$) of the Log-Gabor filter, i.e.
\begin{equation}
\phi_r = \begin{cases}
\pi/2 - \angle_\mathrm{PS} + 2(r-1)\frac{\angle_\mathrm{PS}}{N_r-1},&
\mathrm{for}\ \ r\!>\!1\\
\pi/2,& \mathrm{otherwise}
\end{cases}
\end{equation}
Based on preliminary experiments, viable parameter ranges shown in Table~\ref{tbl:parameterRanges} were used in a grid search. 
We used a single combined metric as optimization objective, namely the \emph{mean metric} given by the mean of conventional metrics RMSE, oHD, and sHD, as these assess various aspects of a delineation.

\begin{table*}[]
\centering
\caption{Tested range of optimized parameters.}
\label{tbl:parameterRanges}
\setlength\tabcolsep{3.pt}
\footnotesize
\begin{tabular}{lcccccccc}
Parameter  & $\lambda_\mathrm{PS}$ & $\angle_\mathrm{PS}$      & $N_r$          & $\sigma_0$   & $\mu$                 & $k_1$                     & $k_2$        & $k_3$             \\ \hline
\rule{0pt}{2.5ex}%
Range & {[}25, 75{]}\,px          & {[}$\pi / 12$, $\pi/3${]} & {[}1,  3{]}  & {[}0.01, 10{]}   & {[}0.1, 5{]}          & {[}0.1, 1{]}              & {[}0.1, 1{]} & {[}0.1, $1000${]} 
\end{tabular}
\end{table*}

We followed a bootstrapping based approach to generate different subsets of samples from diverseUS to run cross-validation for parameter optimization. 
In a pseudo-random manner, we generated 5 different sets $S_i$, such that each sample from diverseUS is left out once.
Then, we ran a 6-fold cross-validation on each $S_i$ for the parameter optimization.
For each parameter setting the average \emph{mean metric} across all crossfolds was computed to identify the best parameter set $s_{i}$ that maximizes the average crossfold performance within $S_i$.

\subsection{Method Standardization}
\label{sec:standardization}

Some methods and bone localization algorithms in the literature, such as PS, do not necessarily aim for a complete and accurate delineation of the bone surface, but rather act as a filter aiming to enhance the visibility of bone surface in an US image.
Hence, their output may additionally contain tissue interfaces and/or they may return multiple-pixel thick bone surfaces, which would both skew the proposed evaluation metrics to their disadvantage.
In our preliminary tests, such simple baseline approaches compared very unfavorably to our proposed methods for the given metrics.
To enable a fair quantitative comparison and to maximize the delineation performance of these algorithms, we introduce and compare two standardization techniques as post-processing steps:
For PS we apply $(\cdot)_{\max}$~\cite{automaticHacihaliloglu2011}, with which the detection with the highest response along each scanline is picked as the bone surface.
We also employ $(\cdot)_\uparrow$ where first a morphological thinning~\cite{thinningLam1992} is applied on any segmentation result, and then for each scanline the deepest (lowermost) bone surface detection is selected as the output delineation~\cite{ozdemir2016graphical}. 
We apply $(\cdot)_\uparrow$ postprocessing to PS and CPS.
For BFG, any bone surface response is always $l$ pixel thick, so for any hairline delineation we simply pick the mid-pixel of BFG output vertically.
In CBG, no post-processing is needed since the bone surface is given implicitly as a thin layer at the interface between the regions with posterior \textbf{T} and \textbf{S} labels.

\section{Results}
\label{sec:evaluations}

\subsection{Implementation}

Computations were performed on an Intel\,i7 4.00GHz CPU with 16\,GB available RAM.  
We refactored our earlier BFG implementation~\cite{ozdemir2016graphical} leading to speed improvements in feature extraction and classifier training. 

From the results in Table~\ref{tbl:paramOptimization}, it can be seen that half of the hyperparameters have the same optimal values across different sets $S_i$.
Hence, CBG hyperparameters are set accordingly as $\lambda_\mathrm{PS}=25$, $\angle_\mathrm{PS}=\pi / 3$, $N_r=3$, $\sigma_0=0.01$, $\mu=5$, $k_1=0.1$, $k_2=0.5$, and $k_3=100$.

Given that the parameter optimization of all methods were conducted on the diverseUS dataset, test images from forearmUS dataset were bilinearly interpolated to match the pixel resolution of diverseUS.
For the quantitative evaluations, we resize the segmentation results to match the initial GS resolution of forearmUS.
This is done by bilinearly resizing the segmentation distance map and thresholding based on the forearmUS GS resolution.

\begin{table}[]
\centering
\caption{For each subset $S_i$, the optimal parameter set ($s_i$) and the average cross-validation scores in [mm] with the proposed CBG.}
\label{tbl:paramOptimization}
\footnotesize
\setlength\tabcolsep{2.pt}
\begin{tabular}{l|cccccccc|cccc}
Set  & $\lambda_\mathrm{PS}$ & $\angle_\mathrm{PS}$ & {$N_r$} & $\sigma_0$ & $\mu$ & $k_1$ & $k_2$ & $k_3$ & RMSE & oHD  & sHD  & {mean} \\ \hline
$S_1$ & 25                    & $\pi$/3              & 3   & 0.01       & 5     & 0.1   & 0.1   & 100   & 0.56 & 1.59 & 3.37 & 1.84 \\
$S_2$ & 25                    & $\pi$/12             & 1   & 0.1        & 5     & 0.1   & 1     & 100   & 0.56 & 1.45 & 3.28 & 1.76 \\
$S_3$ & 25                    & $\pi$/3              & 3   & 0.01       & 5     & 0.1   & 0.5   & 100   & 0.42 & 1.35 & 2.77 & 1.51 \\
$S_4$ & 25                    & $\pi$/3              & 3   & 0.1        & 5     & 0.1   & 1     & 100   & 0.67 & 1.82 & 3.33 & 1.94 \\
$S_5$ & 25                    & $\pi$/12             & 1   & 1          & 5     & 0.1   & 0.5   & 100   & 0.79 & 1.96 & 2.88 & 1.88   
\end{tabular}
\end{table}

\begin{table}[!h]
\centering
\caption{
Delineation performance measured by RMSE, one-way Hausdorff distance (oHD), symmetric Hausdorff distance (sHD), mean scanline error (MSE), miss percentage (MP), mean Euclidean distance (MED), and 95 percentile oHD (oHD$^*$) and sHD (sHD$^*$) on forearmUS dataset. 
UNet (that was trained on another forearm US dataset) and inter-annotator scores reported in~\cite{ciganovic2019deep} are also presented herein as a reference.
}
\label{Table:quantitativeResults}
\setlength\tabcolsep{2.pt}
\footnotesize
\begin{tabular}{|l|rrrr|rrrr|}
\hline
\multicolumn{1}{|r|}{Results$\rightarrow$} & RMSE & oHD& sHD & MSE & MP & MED & oHD$^*$ & sHD$^*$ \\ 
Methods$\downarrow$ & [mm] & [mm] & [mm] & [mm] & [\%] & [mm] & [mm] & [mm] \\ 
\hline
PS$_\uparrow$~\cite{ozdemir2016graphical} &  0.34 & 1.03 & 3.89 & 0.68 & 0.0 & 0.24 & 0.74 & 2.16 \\
\hline
PS$_\mathrm{max}~\cite{automaticHacihaliloglu2011}$ & 3.31 & 6.76 & 13.75 & 8.48 & 0.0 & 2.59 & 6.14 & 13.22 \\ 
\hline
CPS$_\uparrow$~\cite{quader2014confidence,ozdemir2016graphical} & 1.74 & 3.58 & 4.79 & 1.22 & 40.76 & 1.39 & 3.17 & 2.25 \\
\hline
BFG~\cite{ozdemir2016graphical} & 0.56 & 1.47 & 2.31 & 0.68 & 0.02 & 0.42 & 1.14 & 1.76 \\ 
\hline
CBG & 0.28 & 0.75 & 1.78  & 0.39 & 0.10 & 0.20 & 0.56 & 1.26 \\
\hline
\hline
UNet~\cite{ciganovic2019deep} & 0.18 & 0.46 & 1.15 & 0.14 & --- & --- & --- & --- \\
\hline
Inter-Annotator~\cite{ciganovic2019deep} & 0.91 & 2.56 & 2.69 & 0.13 & --- & --- & --- & --- \\
\hline
\end{tabular}
\end{table}

\subsection{Evaluation Results}

We evaluated bone surface delineation performance of the compared methods on the forearmUS dataset.
In Table~\ref{Table:quantitativeResults} we present a quantitative comparison of the proposed CBG with BFG~\cite{ozdemir2016graphical}, PS~\cite{automaticHacihaliloglu2011}, CPS~\cite{quader2014confidence}, and UNet~\cite{ciganovic2019deep}.
PS and CPS required no training and were parameter-optimized on diverseUS. 
BFG and CBG were trained and parameter-optimized with hold-out sets on diverseUS. 
UNet was trained on another dataset of only forearm images.
Considering a fair comparison given a limited training set with potential domain-shift, we compare below the methods that can cope with such practical limitations.

It is seen that PS delineation results can be greatly improved by a simple post-processing step ($\cdot_\uparrow$), over its ($\cdot_{\max}$) alternative used in earlier works~\cite{automaticHacihaliloglu2011}. 
Indeed, for some metrics (RMSE, MED, oHD, and MP), PS$_\uparrow$ even outperforms BFG on the forearm dataset, contrary to earlier findings in~\cite{ozdemir2016graphical}; although BFG still yields a substantially improved ($40\%$) sHD compared to PS$_\uparrow$.
Using only the diverseUS training, our proposed method CBG achieves the best performance in all metrics except MP.
The improvement of CBG over PS$_\uparrow$ for these metrics range from $14\%$ (MED) to $54\%$ (sHD), average being $32\%$.
Similarly, CBG improves our earlier proposed method BFG $42\%$ on average for these metrics.
For all metrics except MSE, both BFG and CBG achieve performance superior to inter-annotator variation reported for this dataset in~\cite{ciganovic2019deep}. 
Since neural networks require a large training set and are susceptible to domain shift, the UNet would have been quite disadvantaged if trained on diverseUS, due to its limited 37 images, among which only few are of the forearm region. 
We therefore presented here UNet results trained with another forearm dataset of 1385 images from a different subject, with which UNet then expectedly outperforms all other methods substantially, in agreement with~\cite{ciganovic2019deep}.
Given that it is not always practical to annotate and train on large sets of a targeted anatomy, UNet results reported in~\cite{ciganovic2019deep} are herein only presented as a reference.
Other works in literature using a similar experiment setup (SonixTouch US machine) report UNet results of 2.43\,mm~\cite{alsinan2019automatic} and 0.39\,mm~\cite{puyang2018simultaneous} for MED metric when only B-mode US images were used as input, where training set sizes were 300 and 415 images, respectively.

\begin{figure*}
\centering
	\begin{tabular}{@{}c@{\hspace{.4em}}c@{\,}c@{\hspace{.2em}}|@{\hspace{.2em}}c@{\,}c@{\hspace{.2em}}|@{\hspace{.2em}}c@{\,}c@{\hspace{.2em}}|@{\hspace{.2em}}c@{\,}c@{\,}c}
    & \multicolumn{2}{c}{RMSE} & \multicolumn{2}{c}{oHD} & \multicolumn{2}{c}{sHD} & \multicolumn{2}{c}{MSE} \\
    & CBG & PS$_\uparrow$ & CBG & PS$_\uparrow$ & CBG & PS$_\uparrow$ & CBG & PS$_\uparrow$ \\[-.3ex]
	\raisebox{.04\textwidth}{\rotatebox[origin=c]{90}{Best}} &
	\includegraphics[width=0.118\textwidth , trim= 1.96cm 1.24cm 1.43cm 0.86cm, clip]{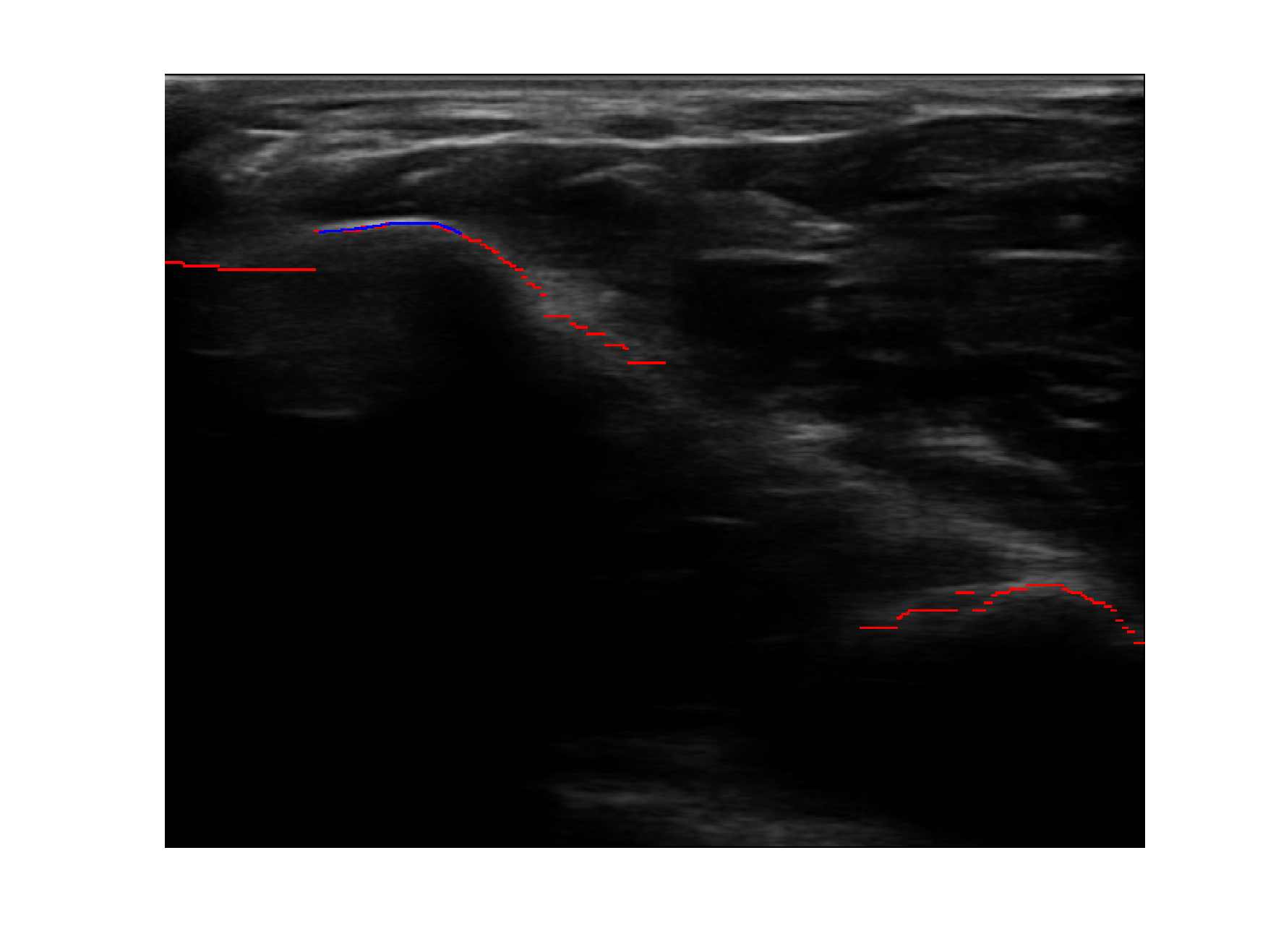} &
	\includegraphics[width=0.118\textwidth , trim= 1.96cm 1.24cm 1.43cm 0.86cm, clip]{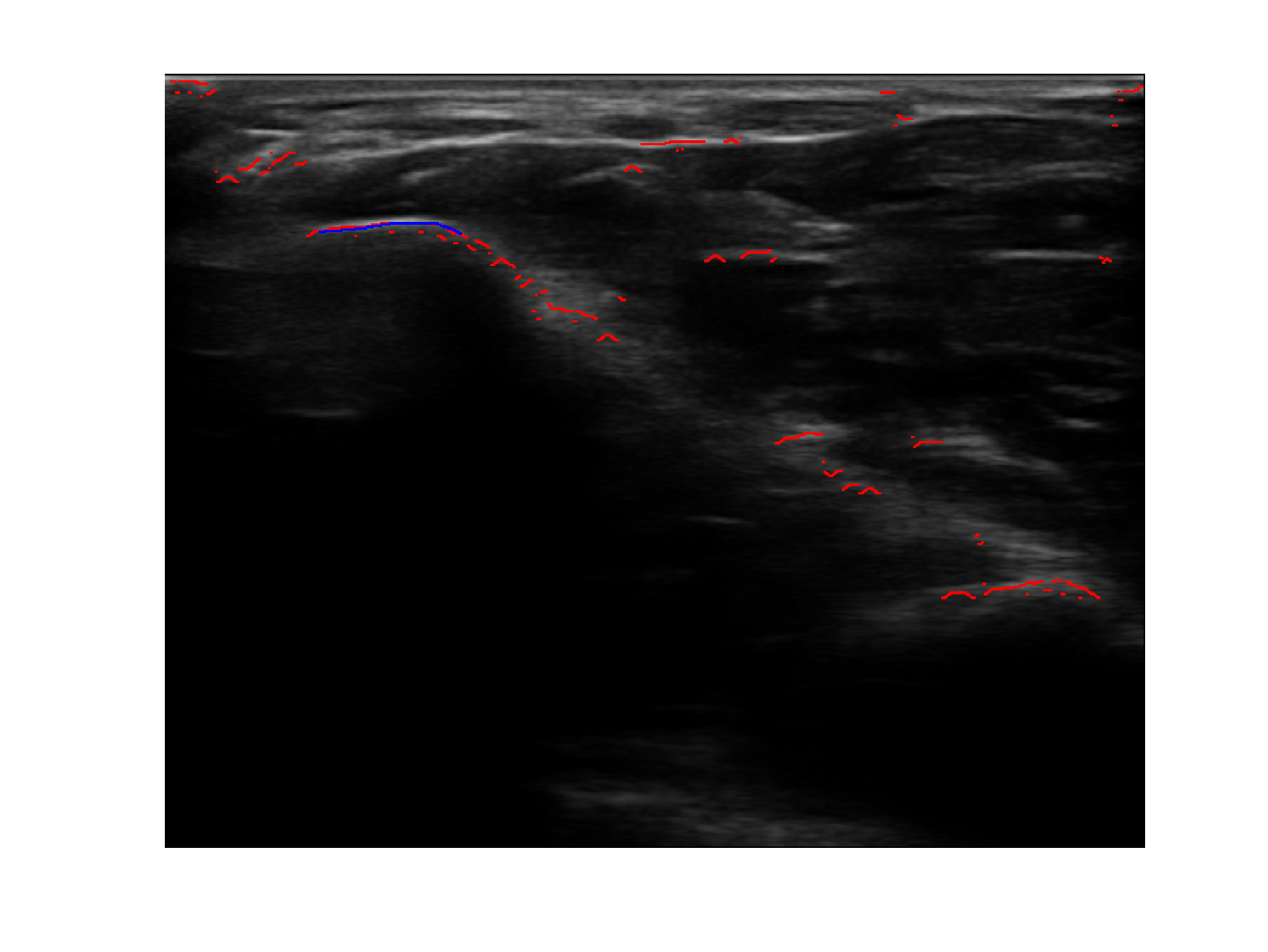} &
    \includegraphics[width=0.118\textwidth , trim= 1.96cm 1.24cm 1.43cm 0.86cm, clip]{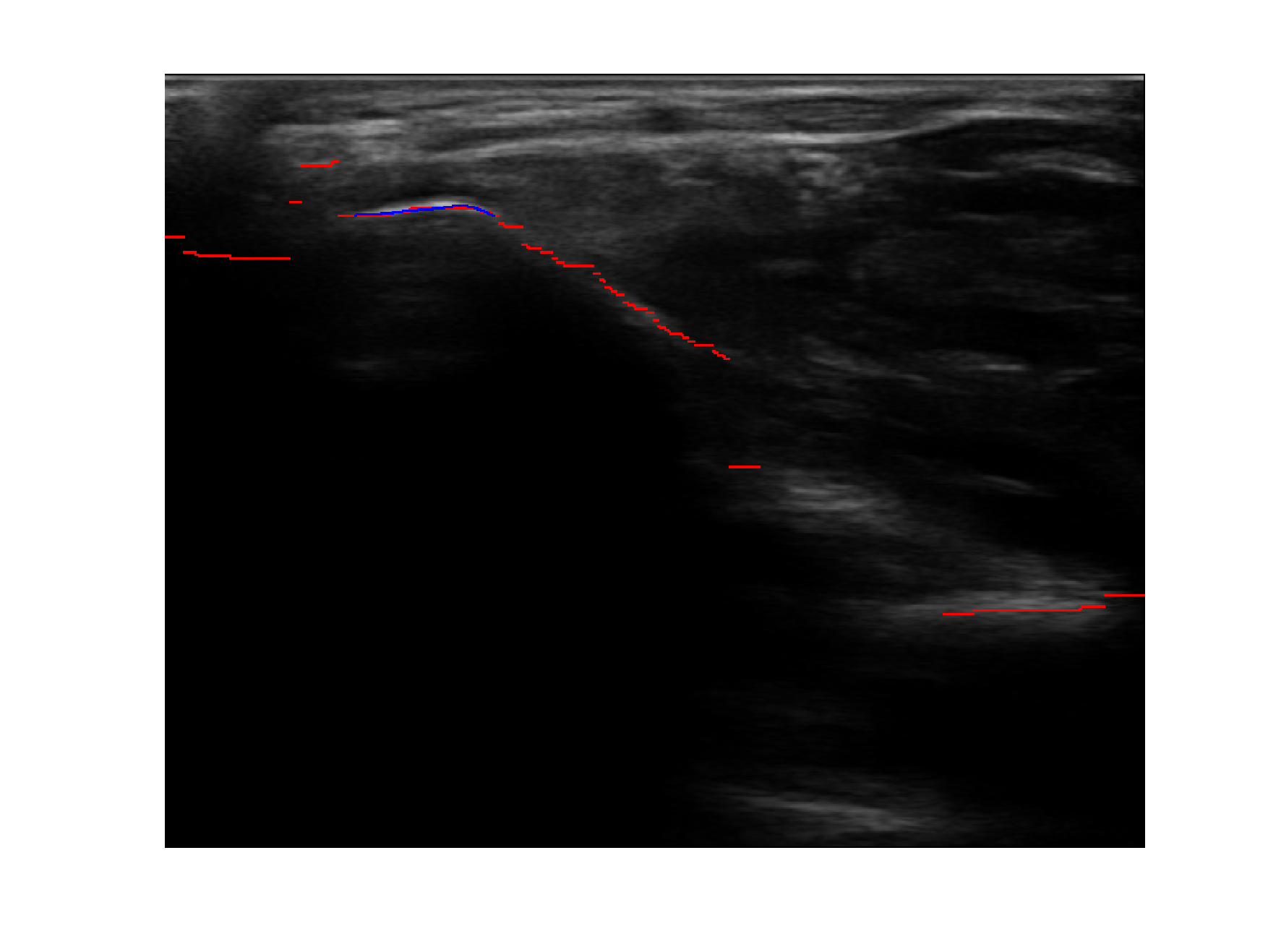} &
    \includegraphics[width=0.118\textwidth , trim= 1.96cm 1.24cm 1.43cm 0.86cm, clip]{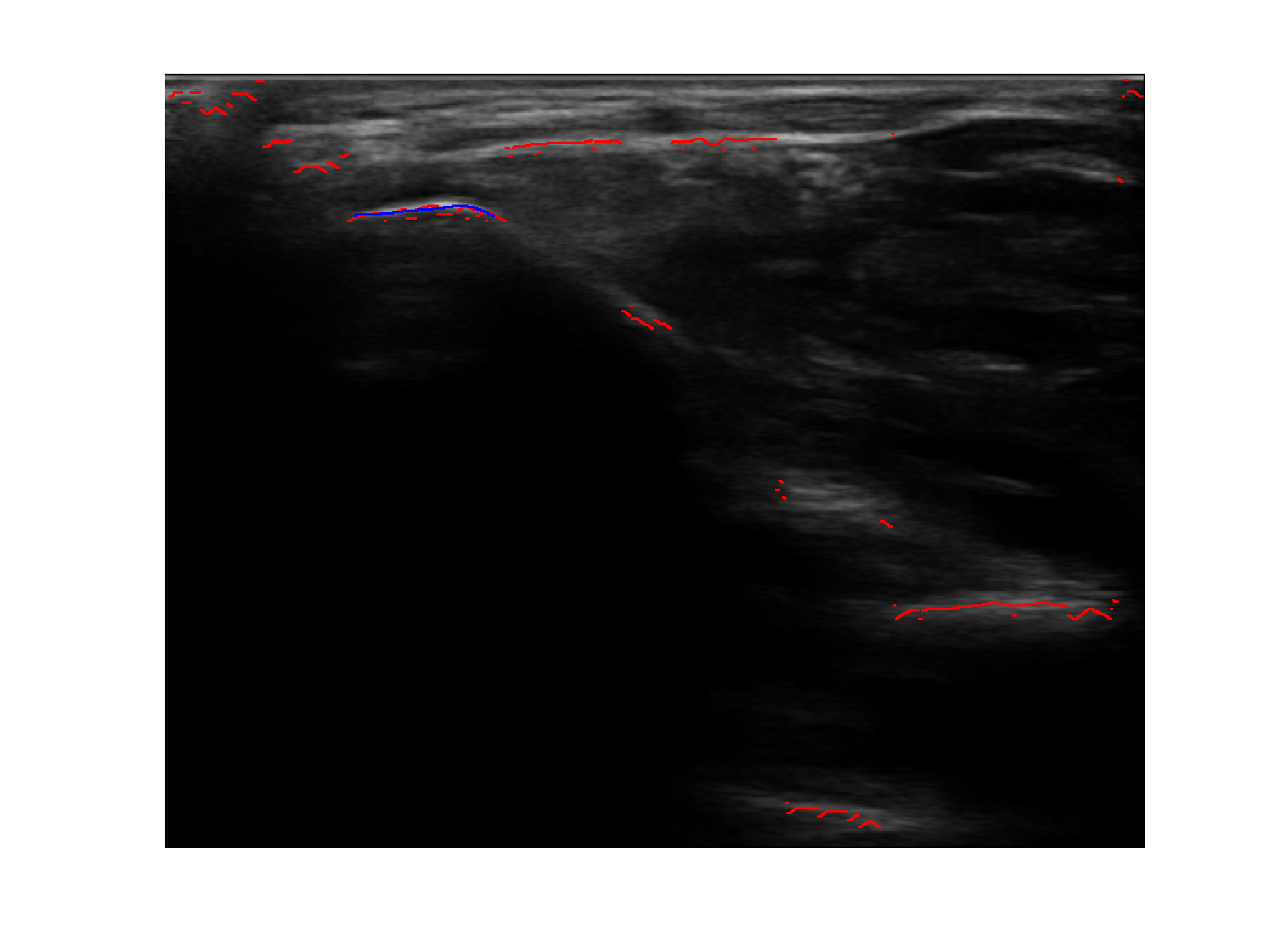} &
    \includegraphics[width=0.118\textwidth , trim= 1.96cm 1.24cm 1.43cm 0.86cm, clip]{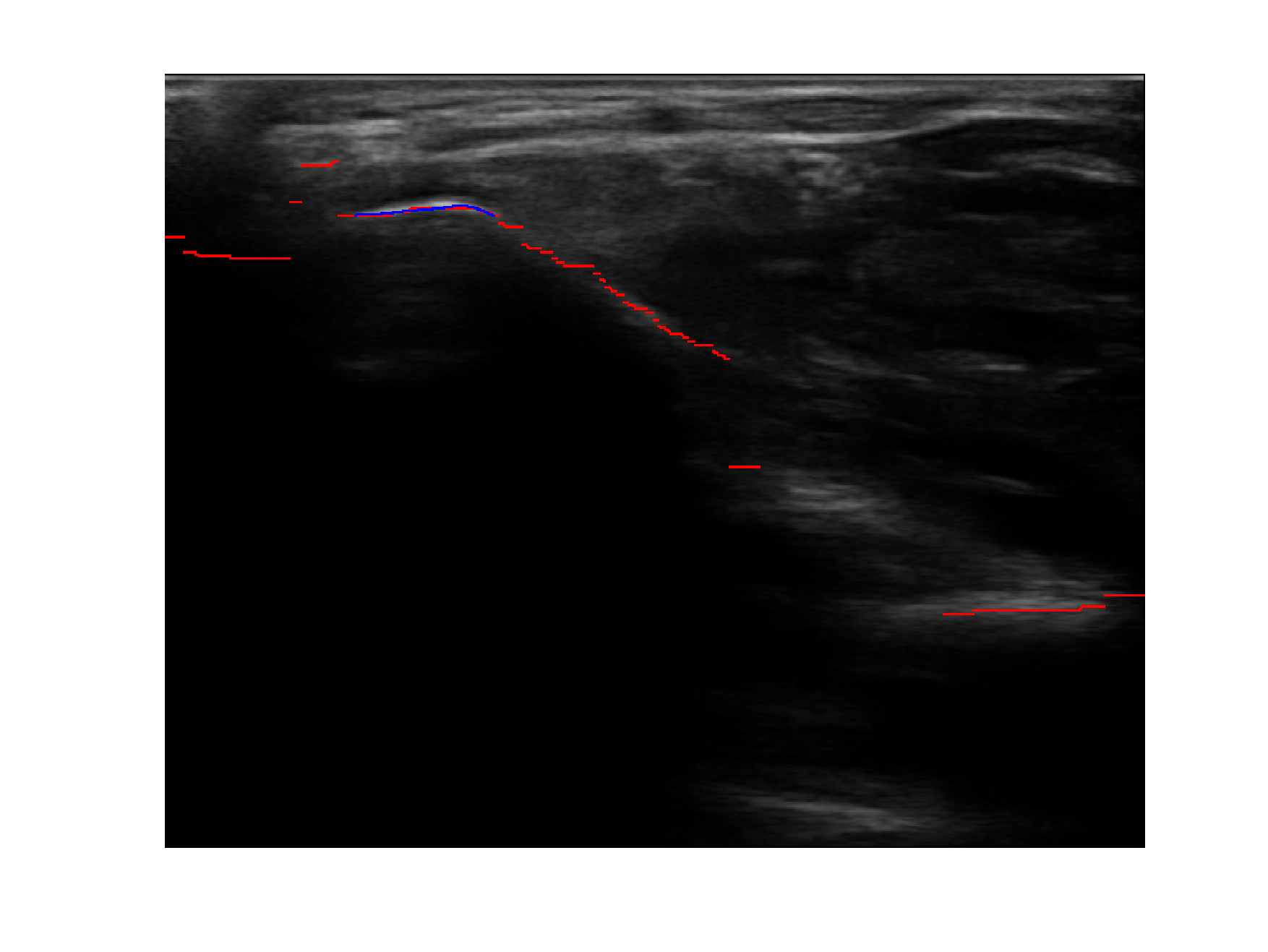} &
    \includegraphics[width=0.118\textwidth , trim= 1.96cm 1.24cm 1.43cm 0.86cm, clip]{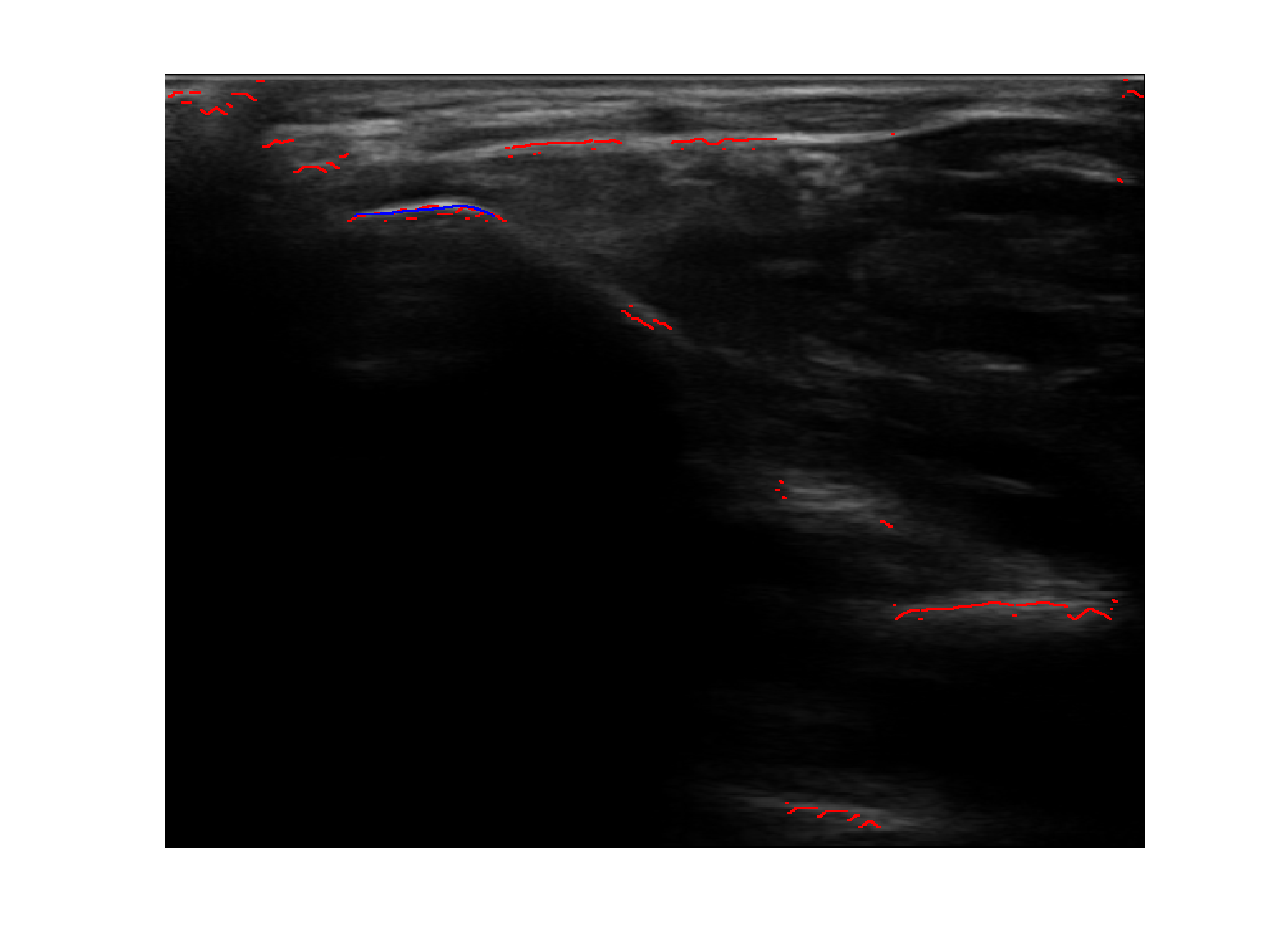} &
	\includegraphics[width=0.118\textwidth , trim= 1.96cm 1.24cm 1.43cm 0.86cm, clip]{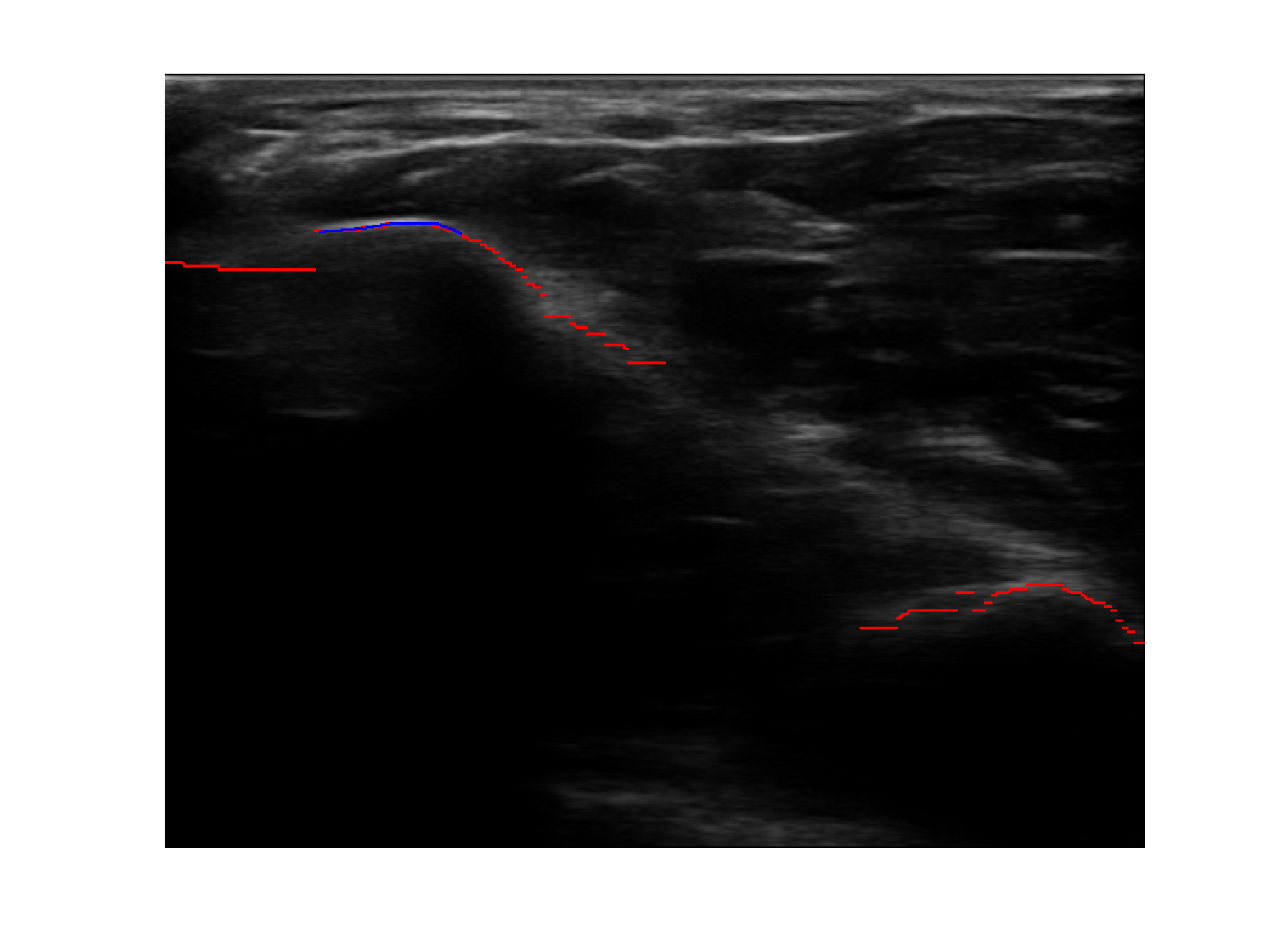} &
	\includegraphics[width=0.118\textwidth , trim= 1.96cm 1.24cm 1.43cm 0.86cm, clip]{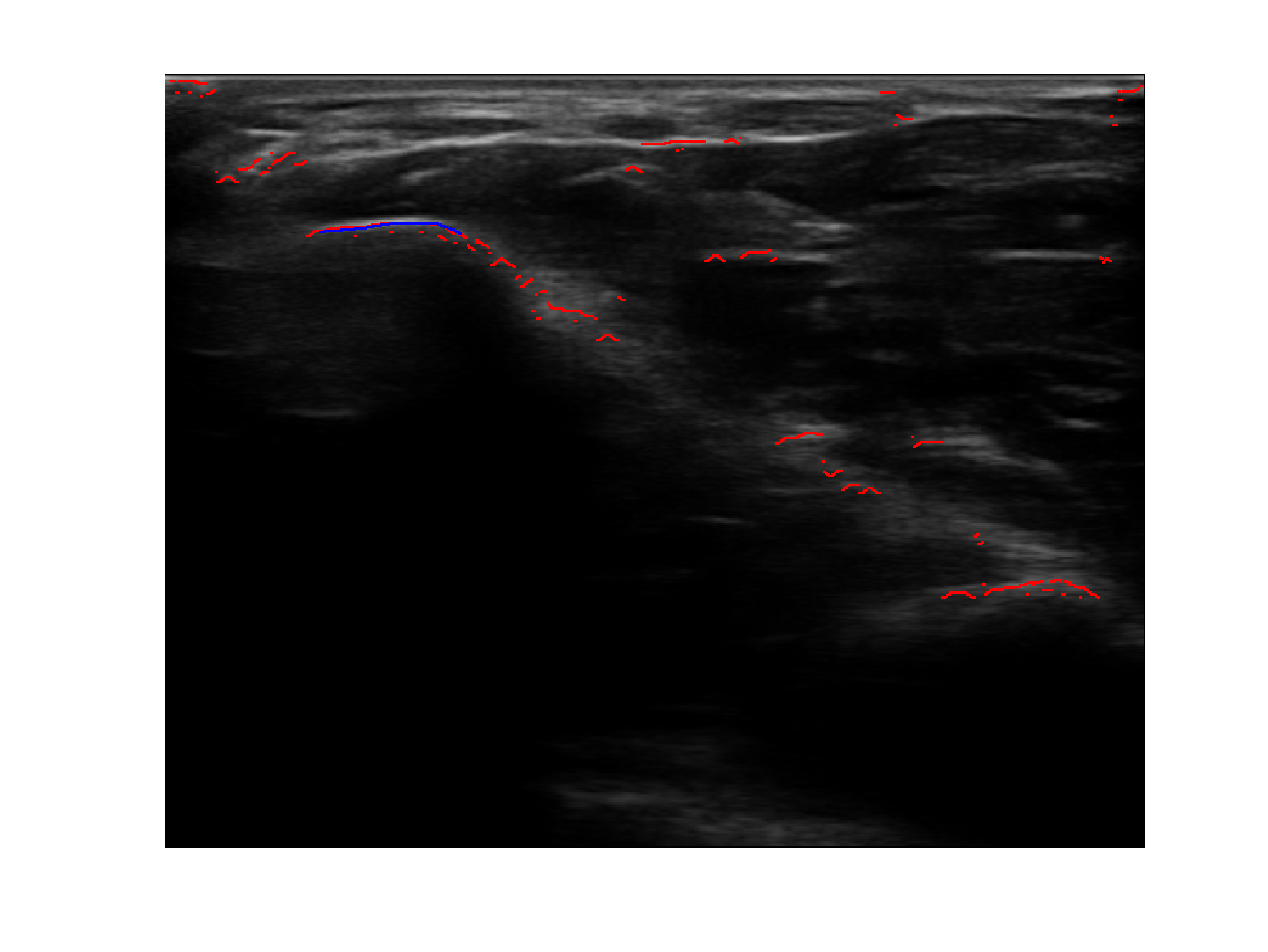}\\ [-.7ex]
	& 0.06 & 0.08 & 0.13 & 0.33 & 0.13 & 0.58 & 0.04 & 0.13 \\
	\raisebox{.04\textwidth}{\rotatebox[origin=c]{90}{Median}} &
    \includegraphics[width=0.118\textwidth , trim= 1.96cm 1.24cm 1.43cm 0.86cm, clip]{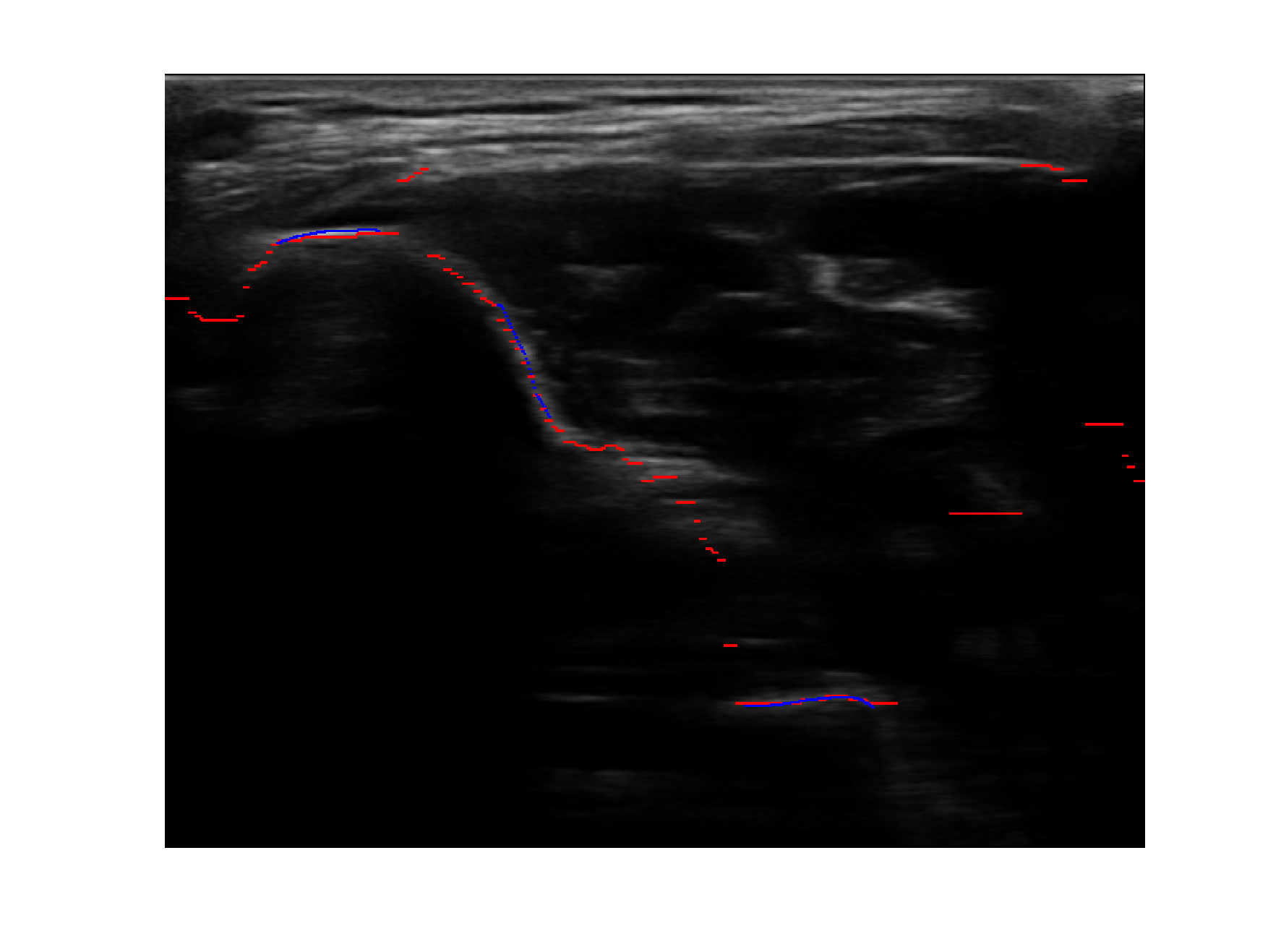} & 
    \includegraphics[width=0.118\textwidth , trim= 1.96cm 1.24cm 1.43cm 0.86cm, clip]{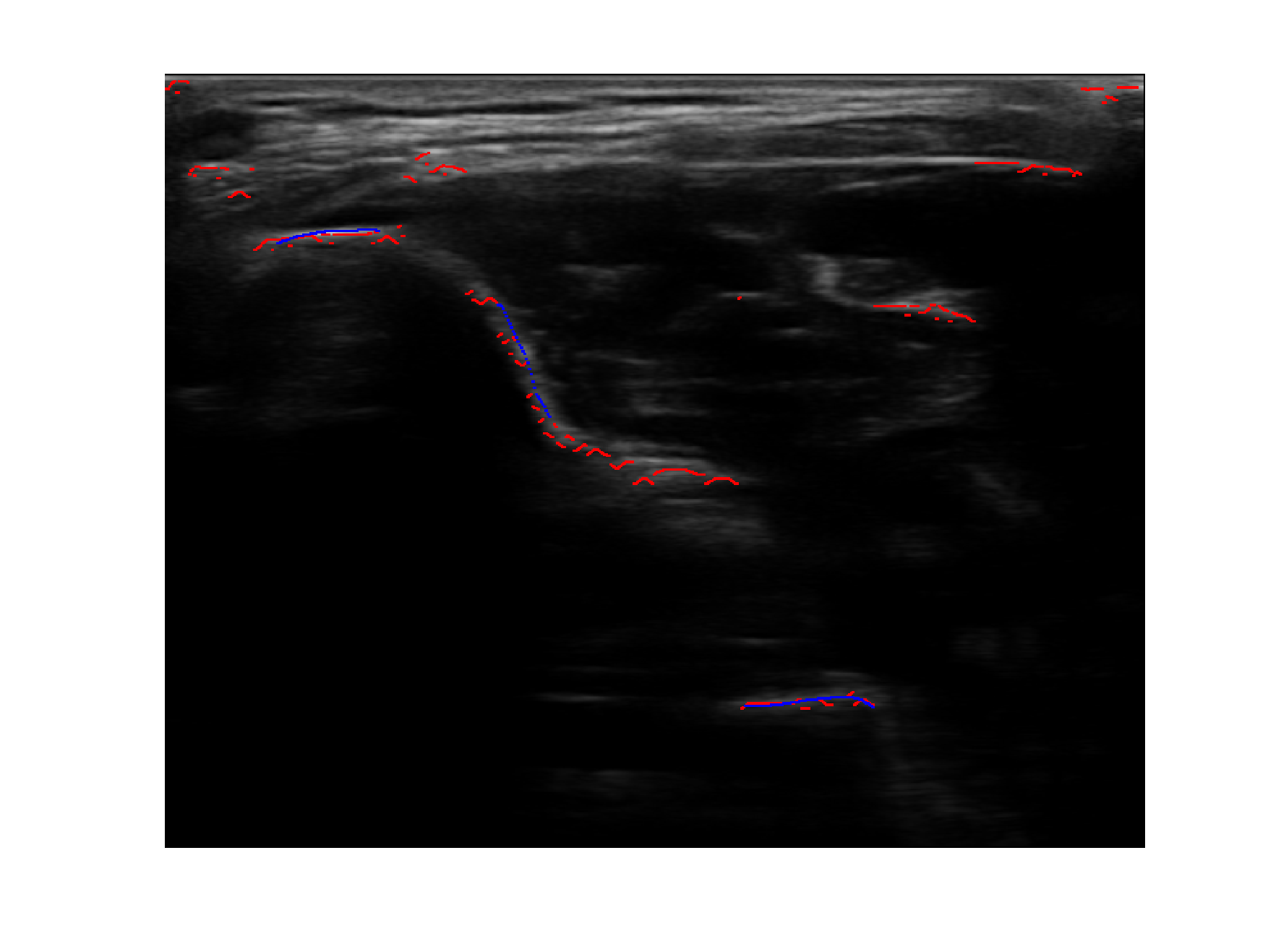} & 
	\includegraphics[width=0.118\textwidth , trim= 1.96cm 1.24cm 1.43cm 0.86cm, clip]{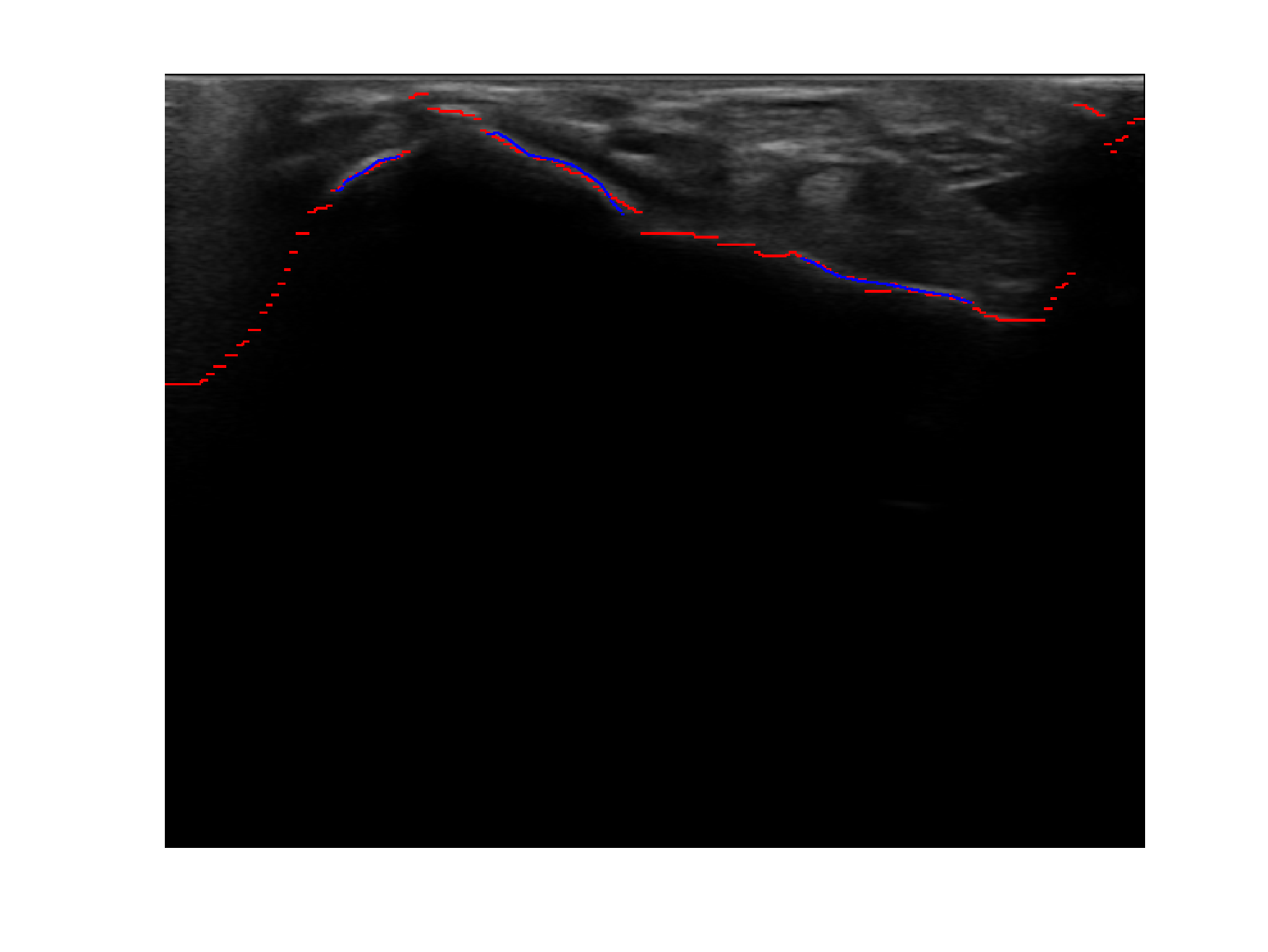} & 
	\includegraphics[width=0.118\textwidth , trim= 1.96cm 1.24cm 1.43cm 0.86cm, clip]{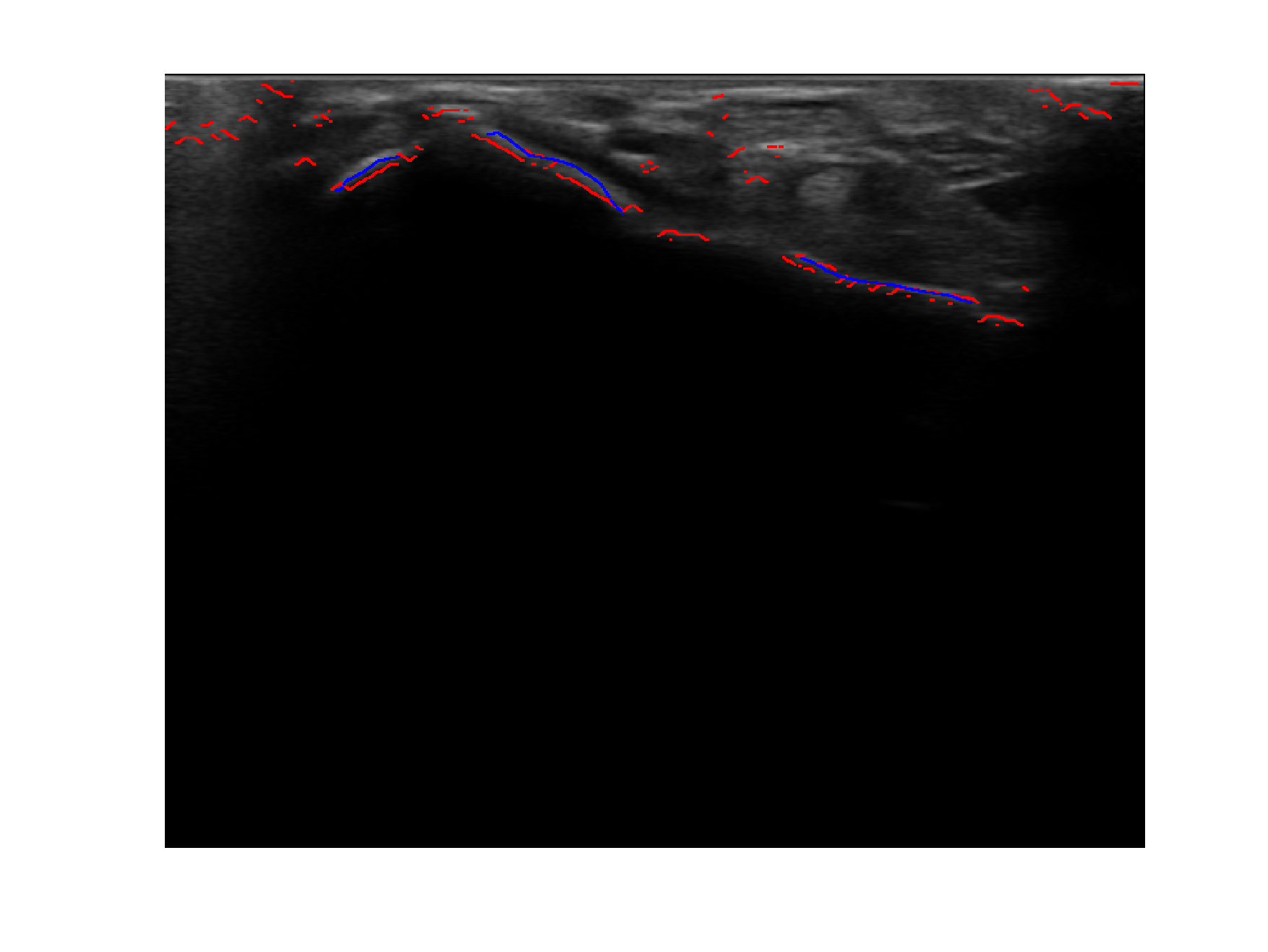} & 
    \includegraphics[width=0.118\textwidth , trim= 1.96cm 1.24cm 1.43cm 0.86cm, clip]{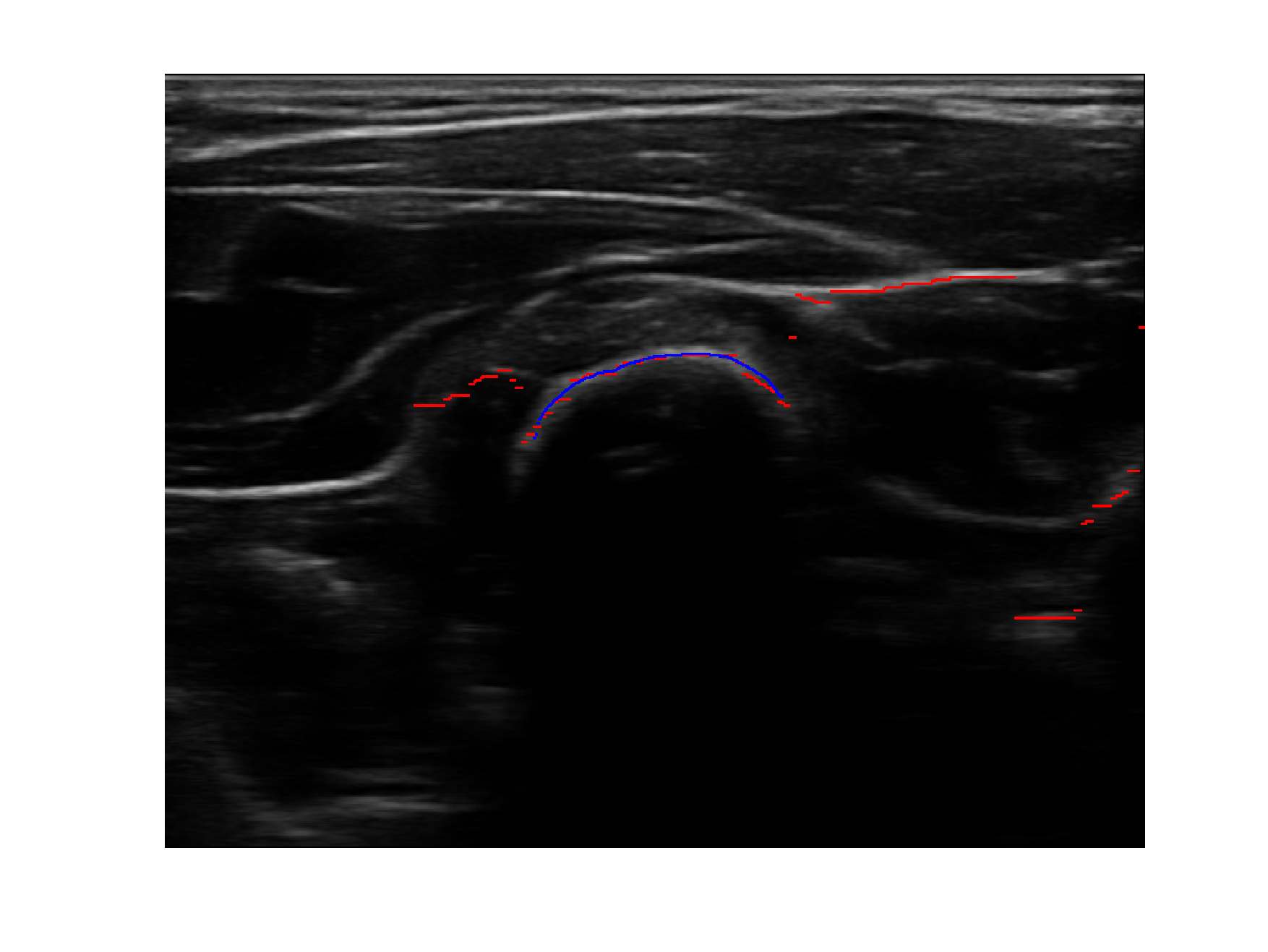} & 
    \includegraphics[width=0.118\textwidth , trim= 1.96cm 1.24cm 1.43cm 0.86cm, clip]{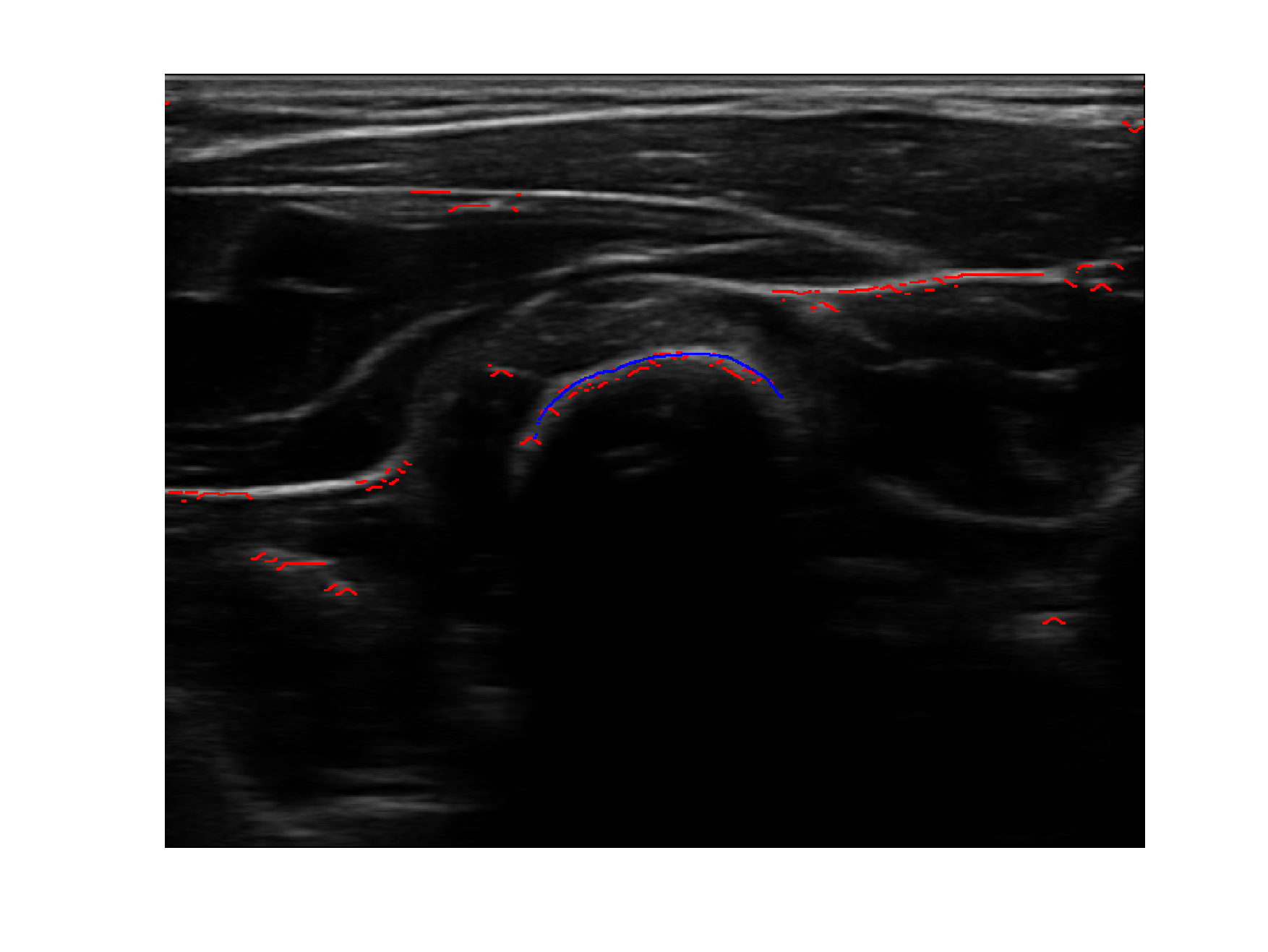} & 
	\includegraphics[width=0.118\textwidth , trim= 1.96cm 1.24cm 1.43cm 0.86cm, clip]{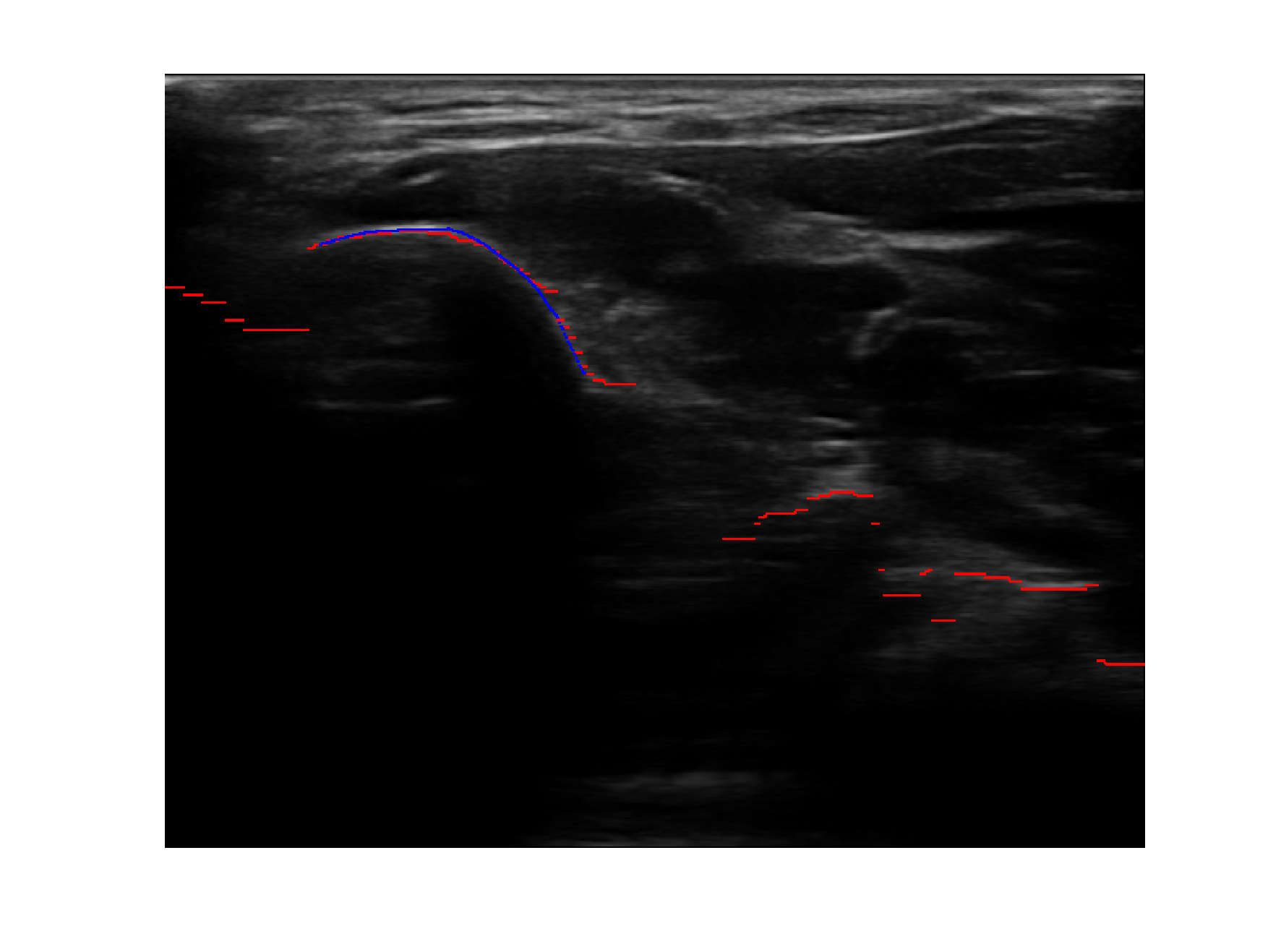} &
	\includegraphics[width=0.118\textwidth , trim= 1.96cm 1.24cm 1.43cm 0.86cm, clip]{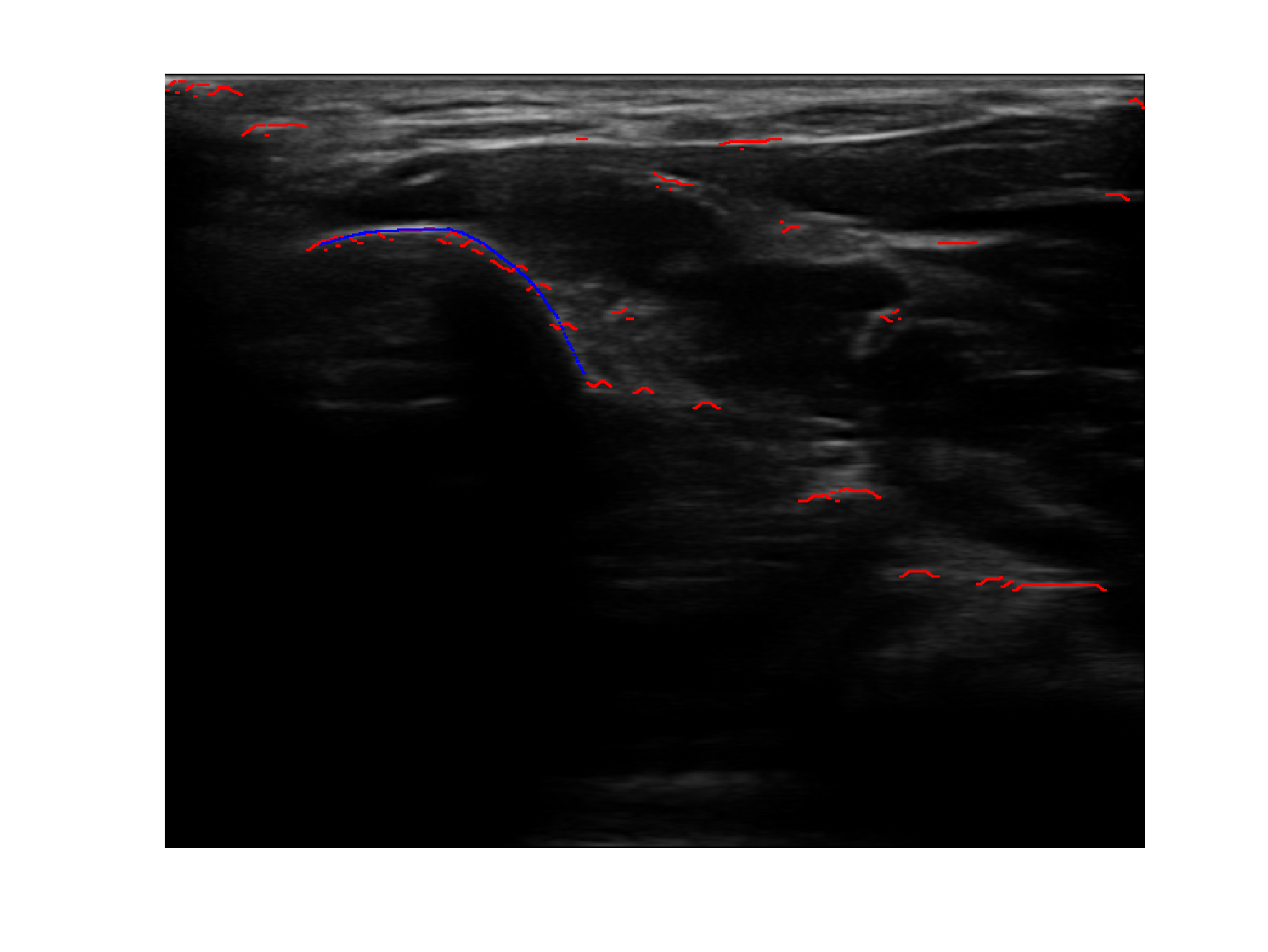} \\[-.7ex]
	& 0.17 & 0.21 & 0.45 & 0.66 & 0.58 & 3.95 & 0.20 & 0.66 \\
	\raisebox{.04\textwidth}{\rotatebox[origin=c]{90}{Worst}} &
    \includegraphics[width=0.118\textwidth , trim= 1.96cm 1.24cm 1.43cm 0.86cm, clip]{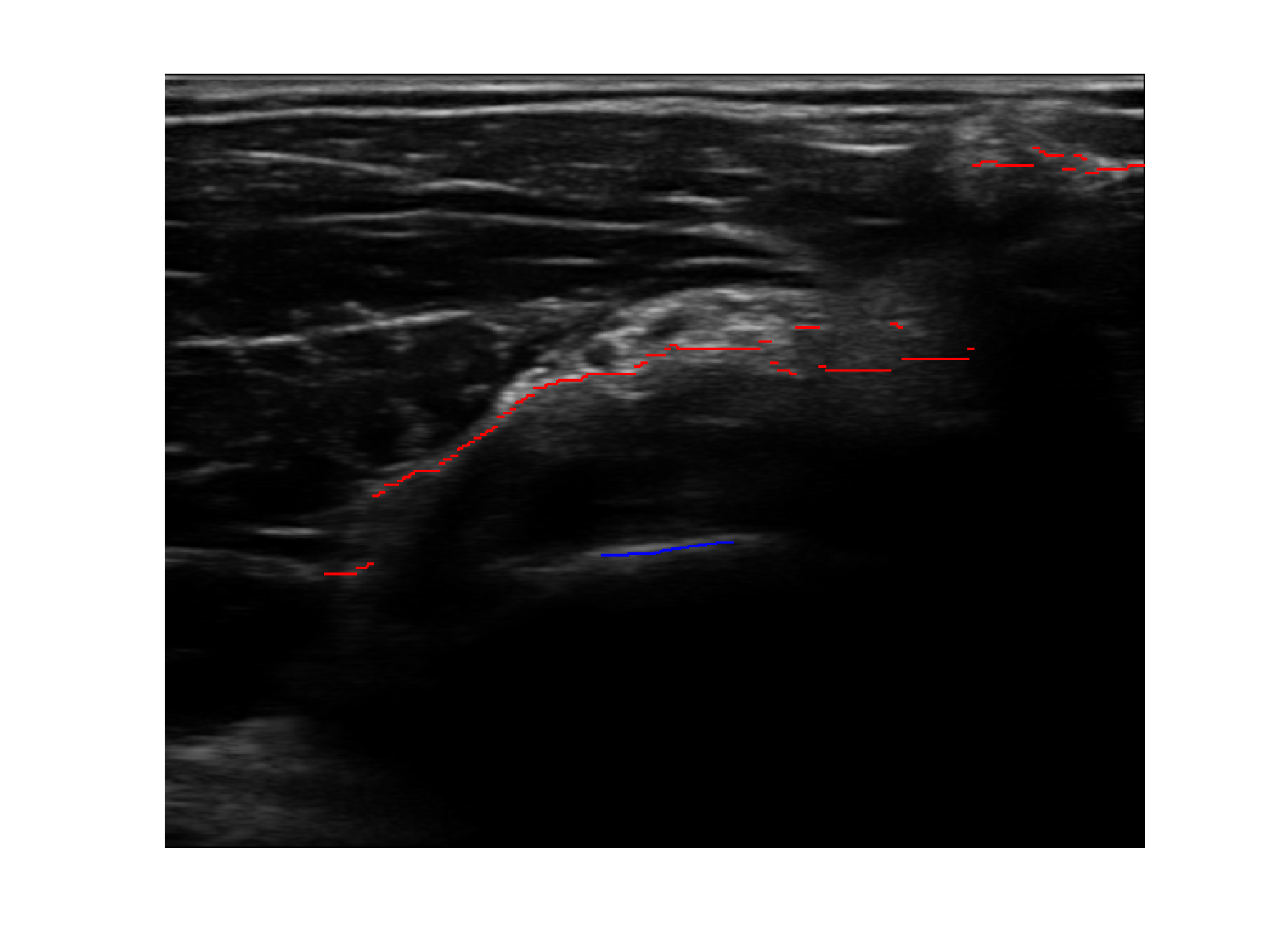} &
    \includegraphics[width=0.118\textwidth , trim= 1.96cm 1.24cm 1.43cm 0.86cm, clip]{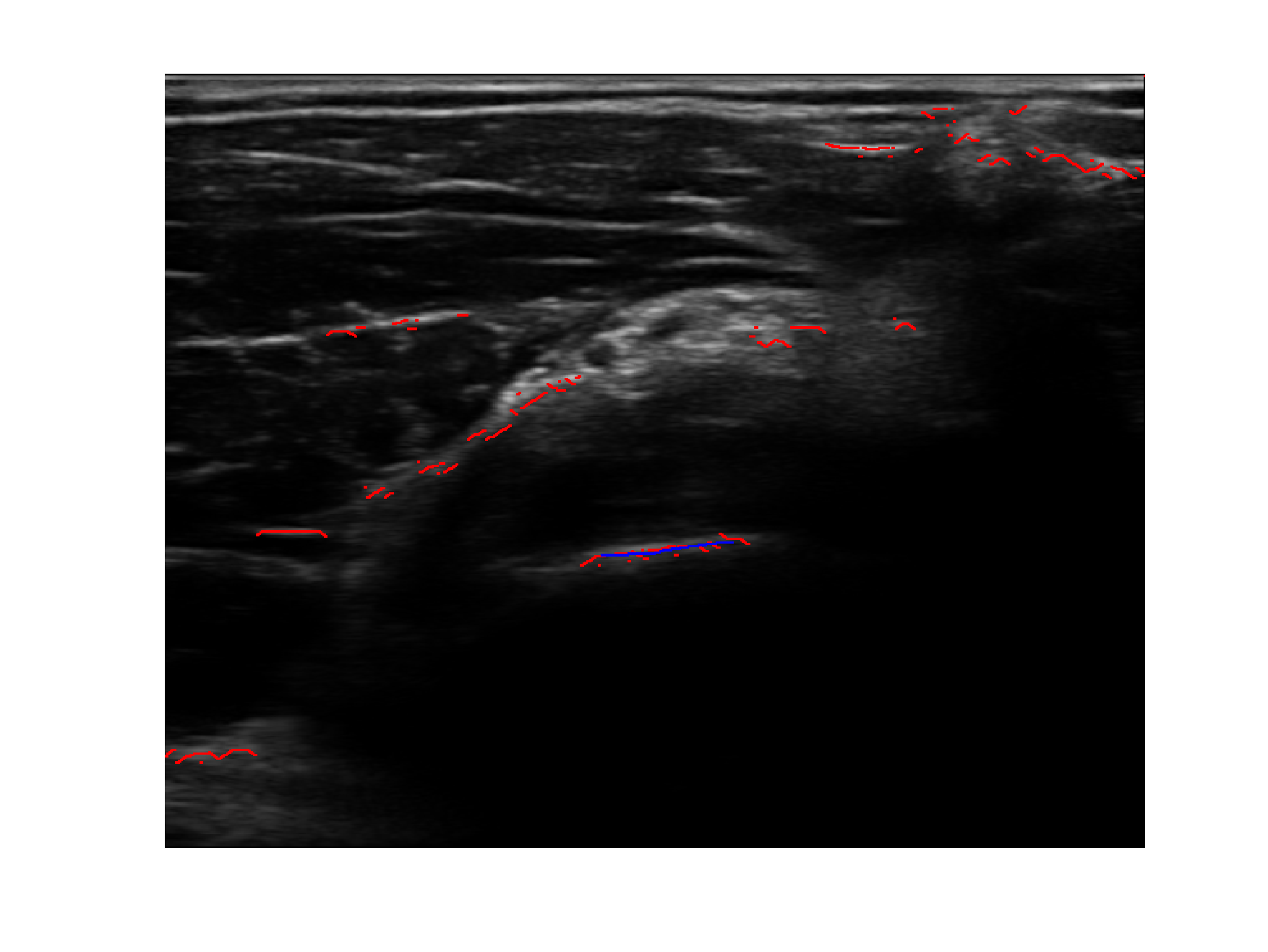} &
	\includegraphics[width=0.118\textwidth , trim= 1.96cm 1.24cm 1.43cm 0.86cm, clip]{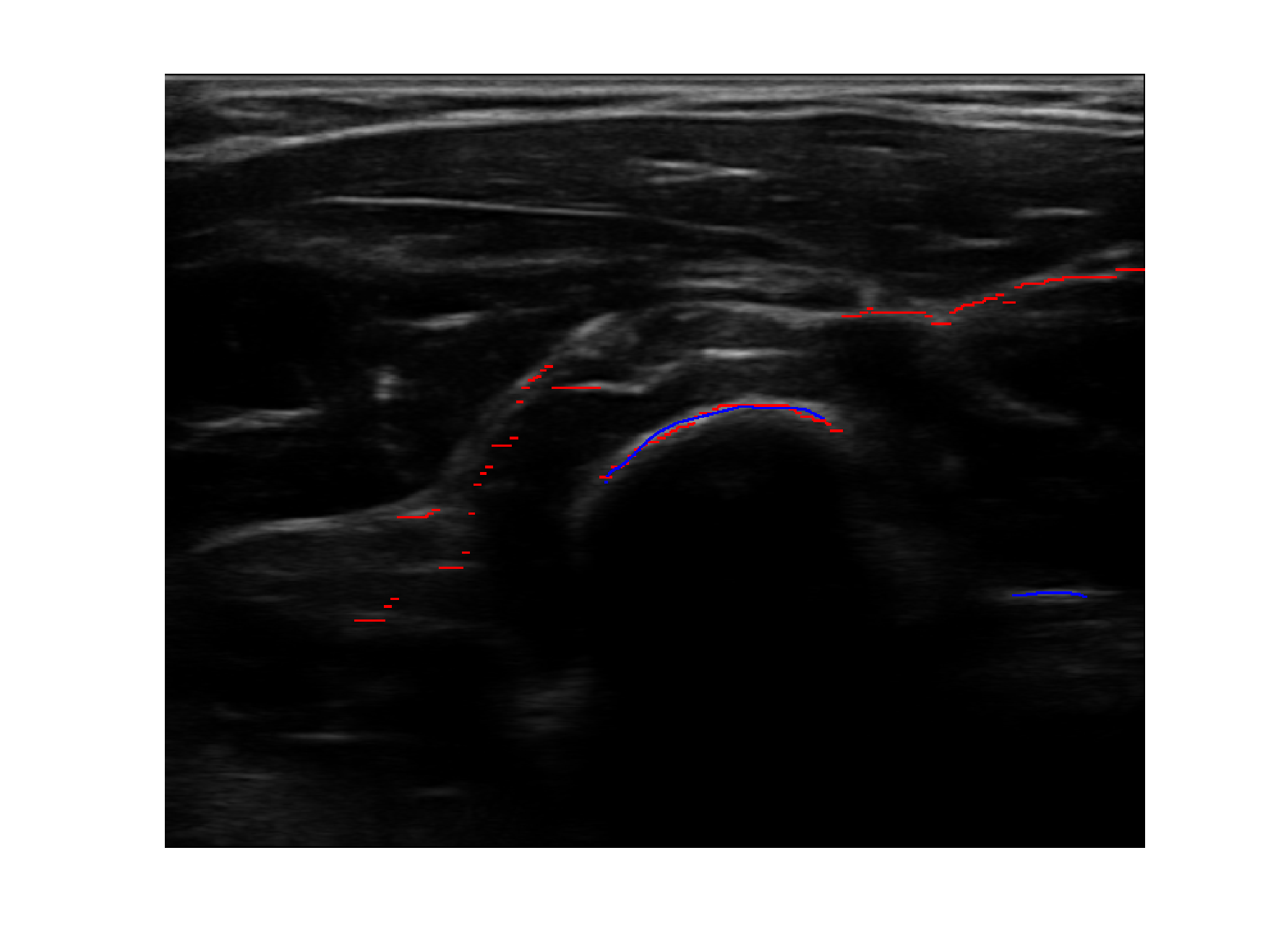} &
	\includegraphics[width=0.118\textwidth , trim= 1.96cm 1.24cm 1.43cm 0.86cm, clip]{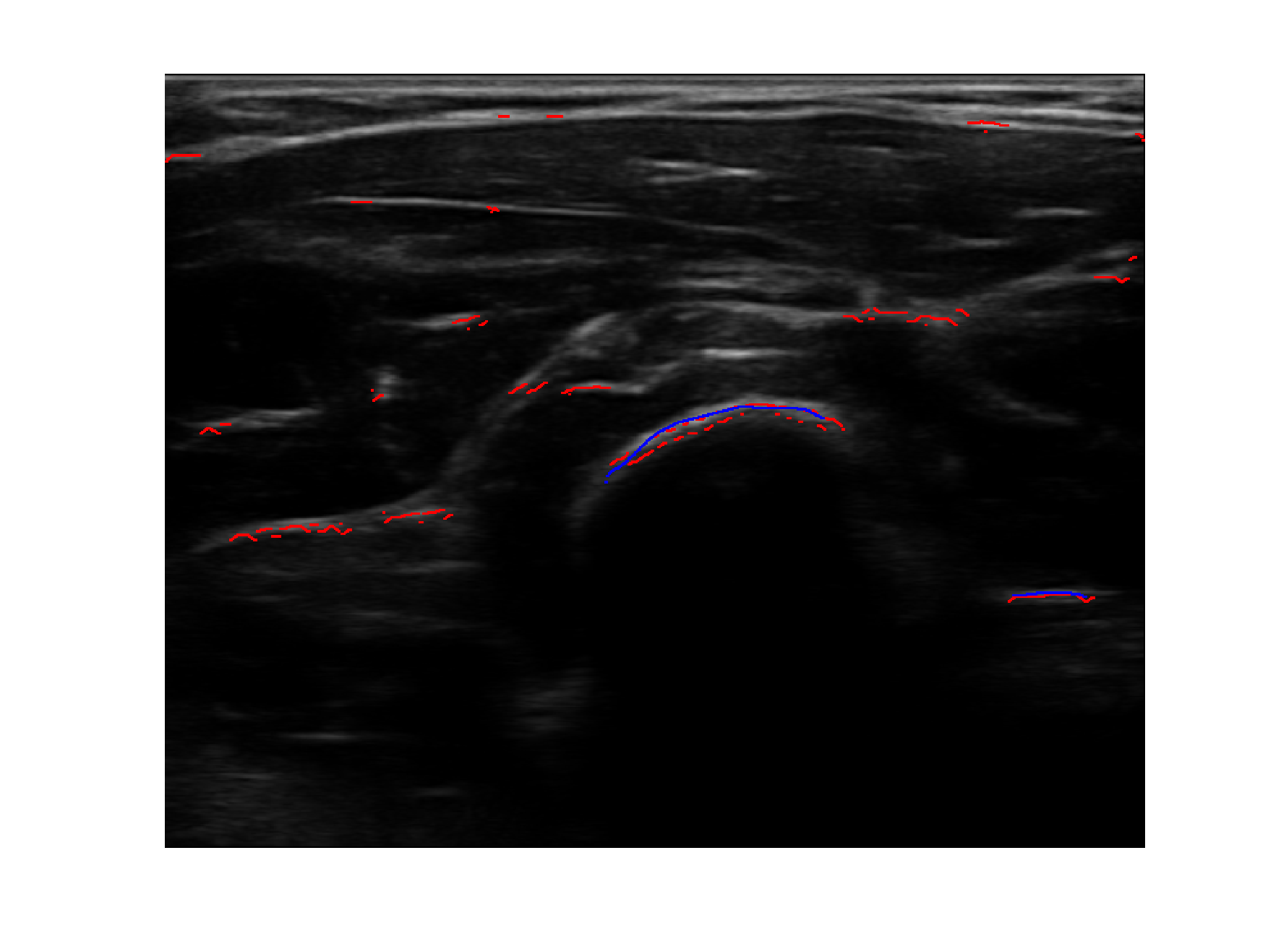} &
    \includegraphics[width=0.118\textwidth , trim= 1.96cm 1.24cm 1.43cm 0.86cm, clip]{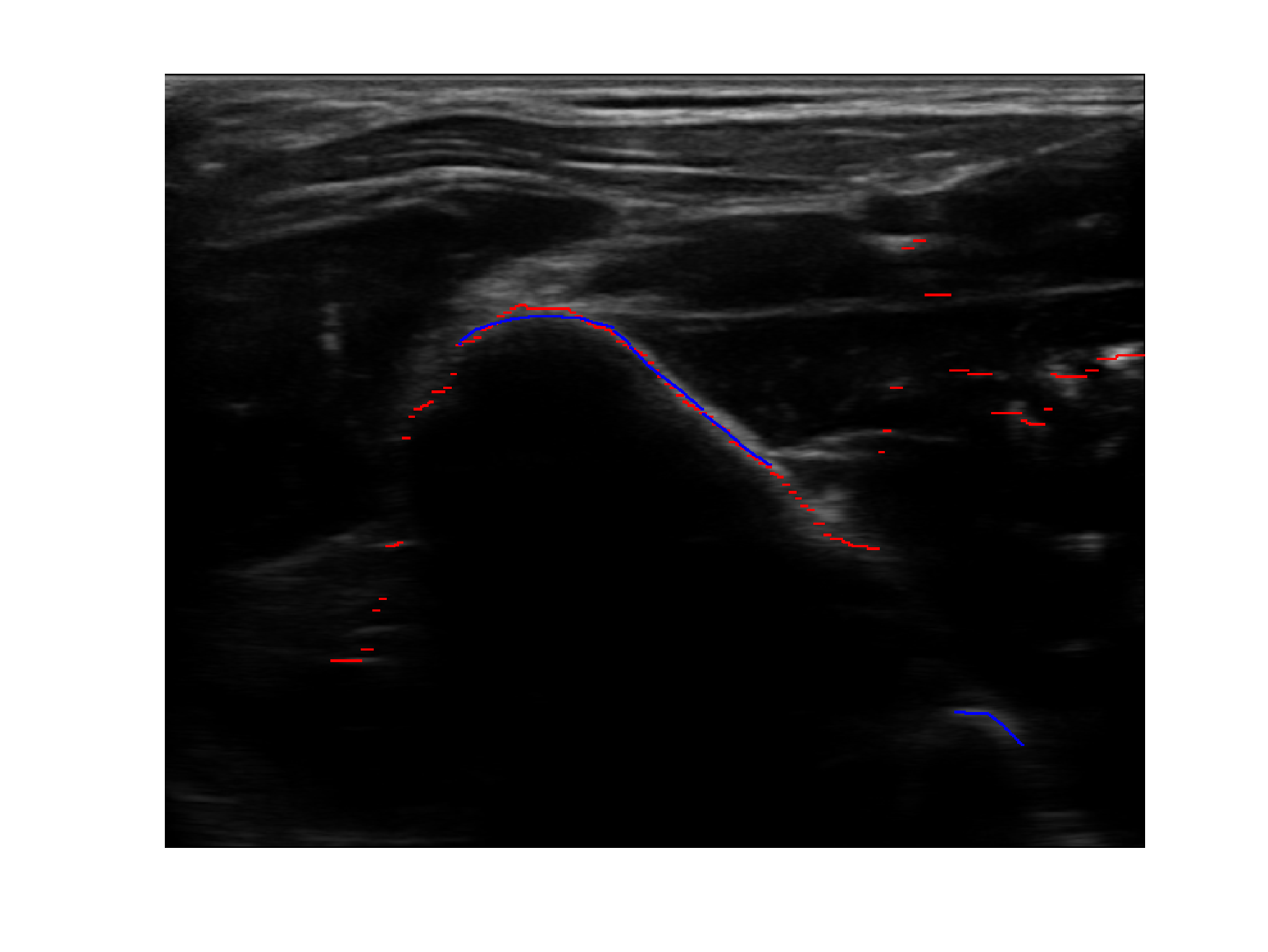} &
    \includegraphics[width=0.118\textwidth , trim= 1.96cm 1.24cm 1.43cm 0.86cm, clip]{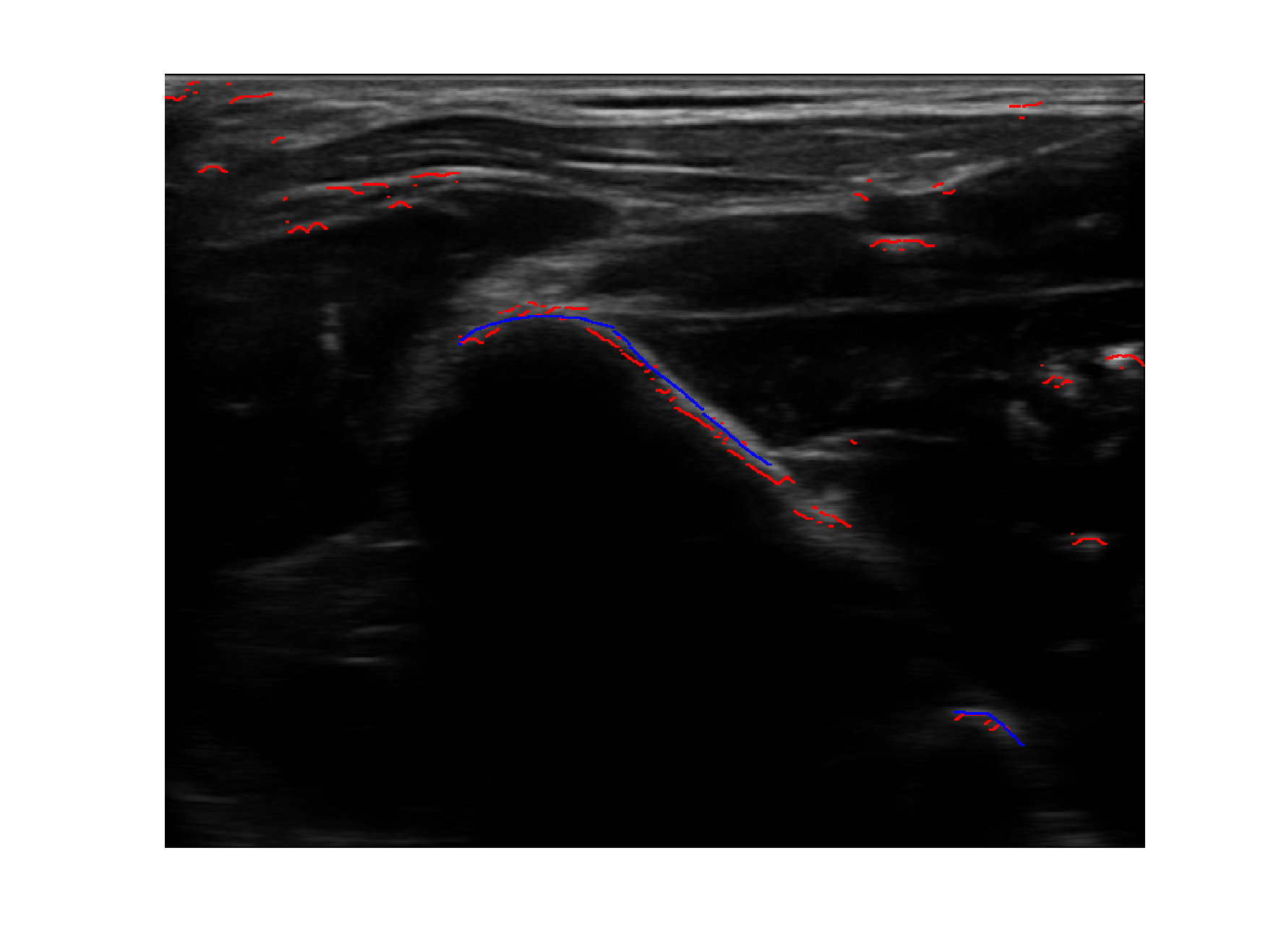} &
	\includegraphics[width=0.118\textwidth , trim= 1.96cm 1.24cm 1.43cm 0.86cm, clip]{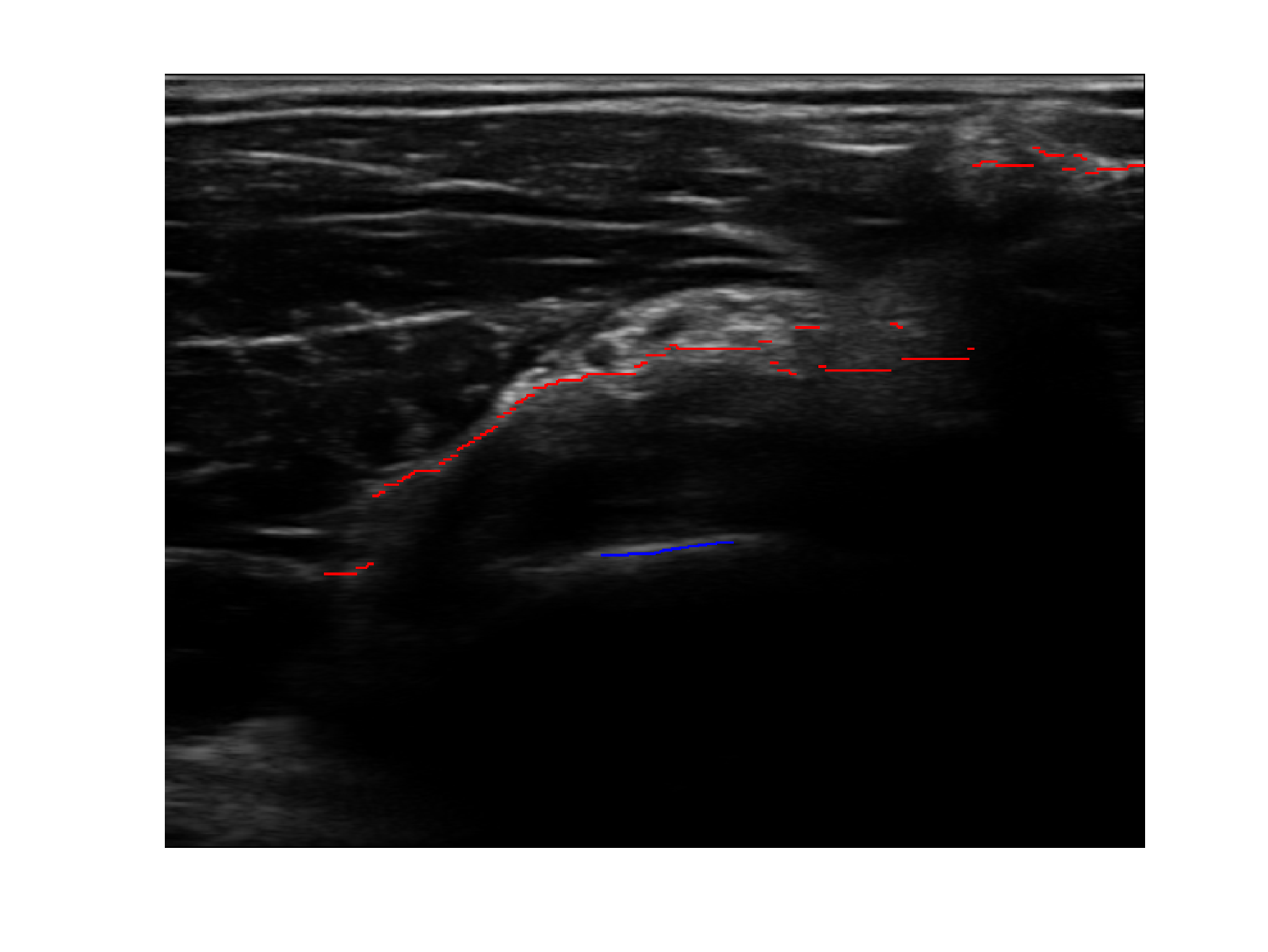} & 
	\includegraphics[width=0.118\textwidth , trim= 1.96cm 1.24cm 1.43cm 0.86cm, clip]{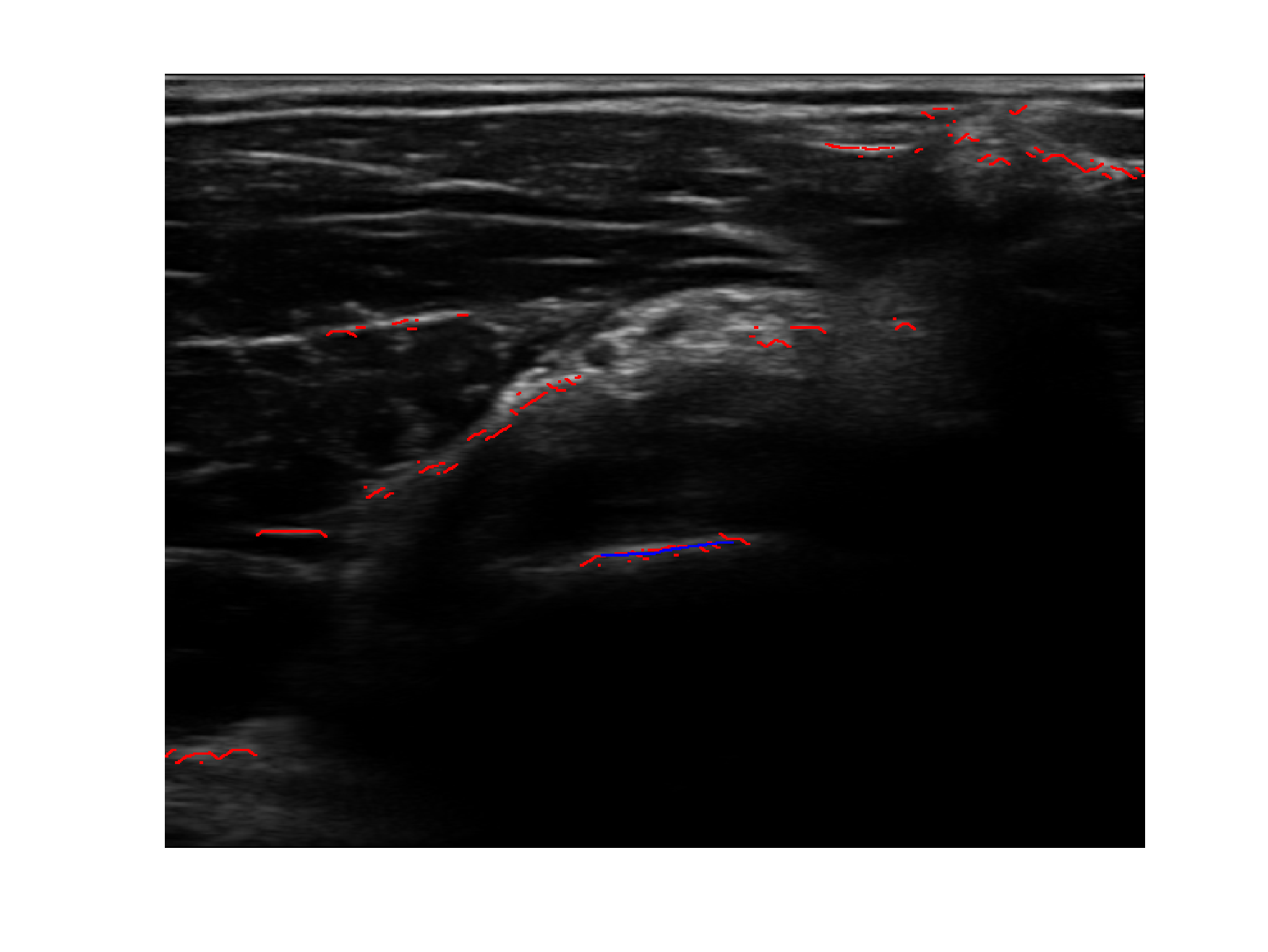}\\[-.7ex]
	& 8.76 & 0.13 & 12.70 & 0.92 & 16.59 & 19.36 & 9.84 & 0.13 \\
    \end{tabular}\\
\caption{
Images yielding the best, median, and worst scores for four main evaluation metrics for our proposed method (CBG) and for PS$_\uparrow$ for comparison.
The red and blue overlays show the method segmentations and the gold standard annotations, respectively.
The quantitative score of each image is displayed below itself.
}
\label{fig:qualitativeResults}
\end{figure*}

In Fig.~\ref{fig:qualitativeResults}, a collection of qualitative results are presented, where the bone surface delineations are demonstrated for images yielding the best, median, and worst scores for four quantitative metrics.
Results for our proposed method CBG are on the odd columns. 
As a comparison and a baseline, bone surface delineation of PS$_\uparrow$ are shown in alternating columns.
It is seen that PS$_\uparrow$ results in many false positive detections, even within the visible bone surface scanlines (GS), which is reflected in its high sHD error.
For instance, PS$_\uparrow$ often detects a bone surface per scanline regardless the existence of a real bone, mainly due to hyperechoic bands within the soft tissue, e.g.,\ subcutaneous tissue.
Such false detections can be mostly avoided in BFG and CBG since to similar patterns are commonly visible and hence can be learned by the classifiers from diverseUS.
Another common pitfall observed in PS$_\uparrow$ is that detections can bounce abruptly between deep and shallow regions.
While there can be multiple bone tissues as well as gaps in between them in a given US image that justify such jumps, a continuous surface is often expected throughout a single bone surface. 
Both BFG and CBG can often avoid such false detections thanks to globally optimized factor graph constraints. 
It is also seen in Fig.~\ref{fig:qualitativeResults} first row that despite appearing very similar to bone surface, the connective tissue between two bones (radius and ulna) is often correctly identified as background in CBG thanks to graphical formulation.
Upon inspection of the worst performing image samples, one can notice that false negatives for CBG occur when there is shadow appearing region with a smooth and continuous hyperechoic band above; e.g.,\ first and last columns in Fig.~\ref{fig:qualitativeResults}. 
Based on visual inspection, metrics sHD and MSE were found to be more representative of perceived delineation quality, whereas oHD was not very correlated with visually pleasing bone surface delineations.

\section{Conclusions}

Earlier works on bone surface localization in US images such as using maximum phase symmetry (PS)~\cite{automaticHacihaliloglu2011} or intensity gradient~\cite{jain2004understanding} often required many assumptions and manual interactions.
Deep learning based techniques have already shown strong improvements in bone surface delineation problem.
However, such methods require large annotated datasets, which are often not available for a targeted anatomical structure. 
For automatic bone surface delineations, we have herein proposed a method to incorporate US physics as constraints into image analysis.
We have introduced a graph formalization, in which the US interaction with bone surfaces and resulting shadows are encoded as graph edge potentials. 
We have used an ensemble training method to compute unary potentials of labels for soft tissue and acoustic shadow.
The transition between the two posterior labels localize the bone surface in a given image.
In the future we aim to formulate our US physics constraints within the scope of recurrent neural networks. 
We believe imaging physics shall be incorporated more in image analysis tasks, and this work to be a step in that direction.

\section{Acknowledgements}

We thank Dr.\,Andreas Schweizer for annotations, and the Swiss National Science Foundation and a Highly Specialized Medicine grant from Zurich Department of Health for funding.

\bibliographystyle{ieeetr}
\bibliography{bib}

\clearpage

\section{Appendix}

\subsection{Feature Selection}
\label{sec:featureSelection}

Some of the features proposed earlier in Section~\ref{sec:features} may not be necessary for a particular classification task, or when used together, some features may not bring significant gain to the learned model.
This is not favourable due to additional computation time needed for each feature. 
There are various methods for feature selection for learning algorithms.
Trying every feature subset combination is intractable, due to the large number of possible combinations ($2^{N_f}$$-$$1$ for $N_f$ features).
A common method is the \emph{greedy backward elimination}, where the feature with the least influence on the trained model is removed from the feature space one at a time and a new model is trained, sequentially eliminating features until a single feature is left or another stopping criterion {(e.g.,\ a large performance drop)} is reached.
Out-of-bag (OOB) error of a classifier is a common metric for estimating the influence of each feature on the classification task.

Since some of our features are extracted efficiently at multiple scales at once, we group all scales of a given feature in the feature selection and determine their group performance as the \emph{maximum} OOB importance in that feature group, which is called the \emph{group OOB importance}.
In each iteration of our feature elimination process, the remaining features (groups) are used to train both shadow and tissue classifiers on a chosen training set $S_i$.
Then, the feature group with the lowest group OOB importance is determined separately for each classifier for elimination from that classifier in the next iteration.

\begin{figure}
\centering
\subfigure{\includegraphics[trim = 15mm 0mm 20mm 0mm, clip,width=0.9\linewidth]{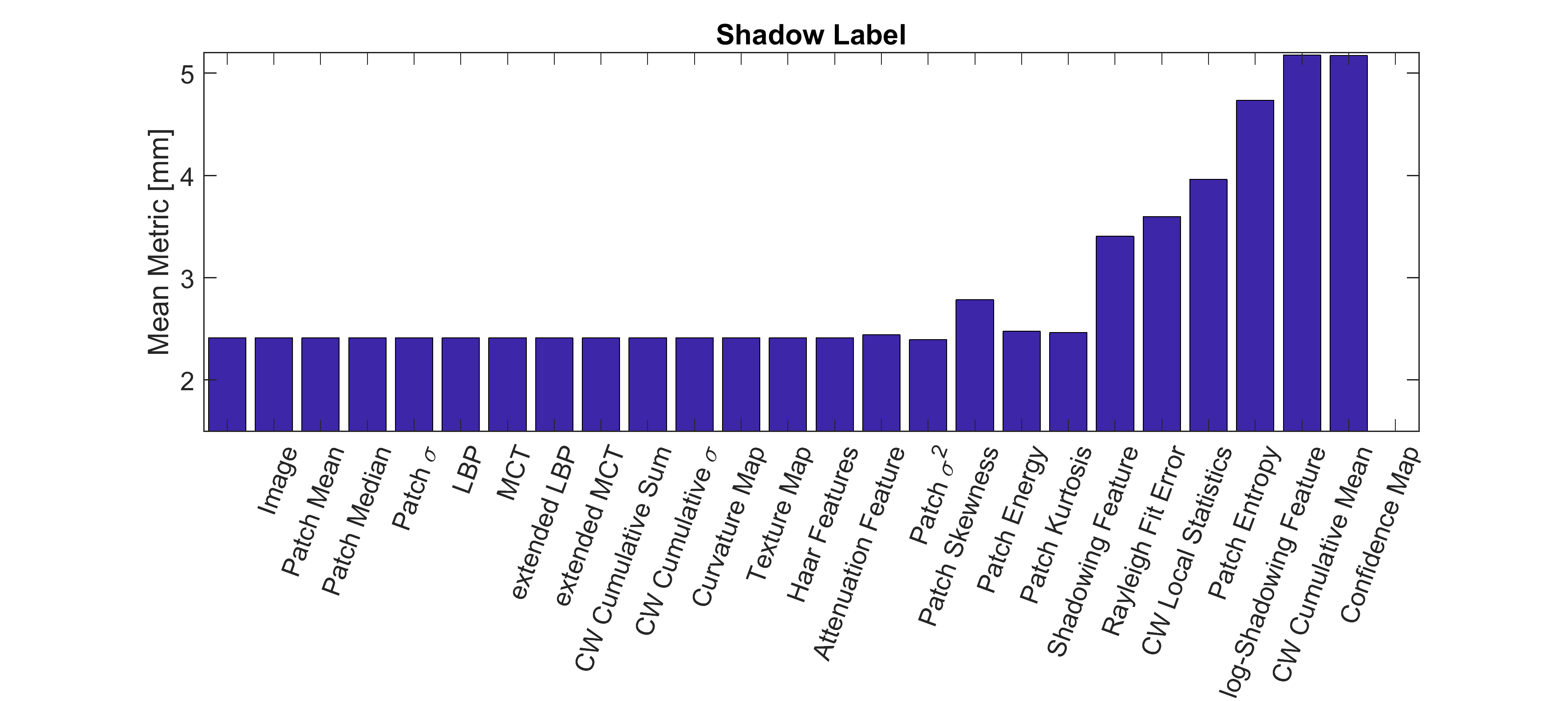}
}%
\\%
\vspace{-1.em}%
\subfigure{\includegraphics[trim = 15mm 0mm 20mm 0mm, clip,width=0.9\linewidth]{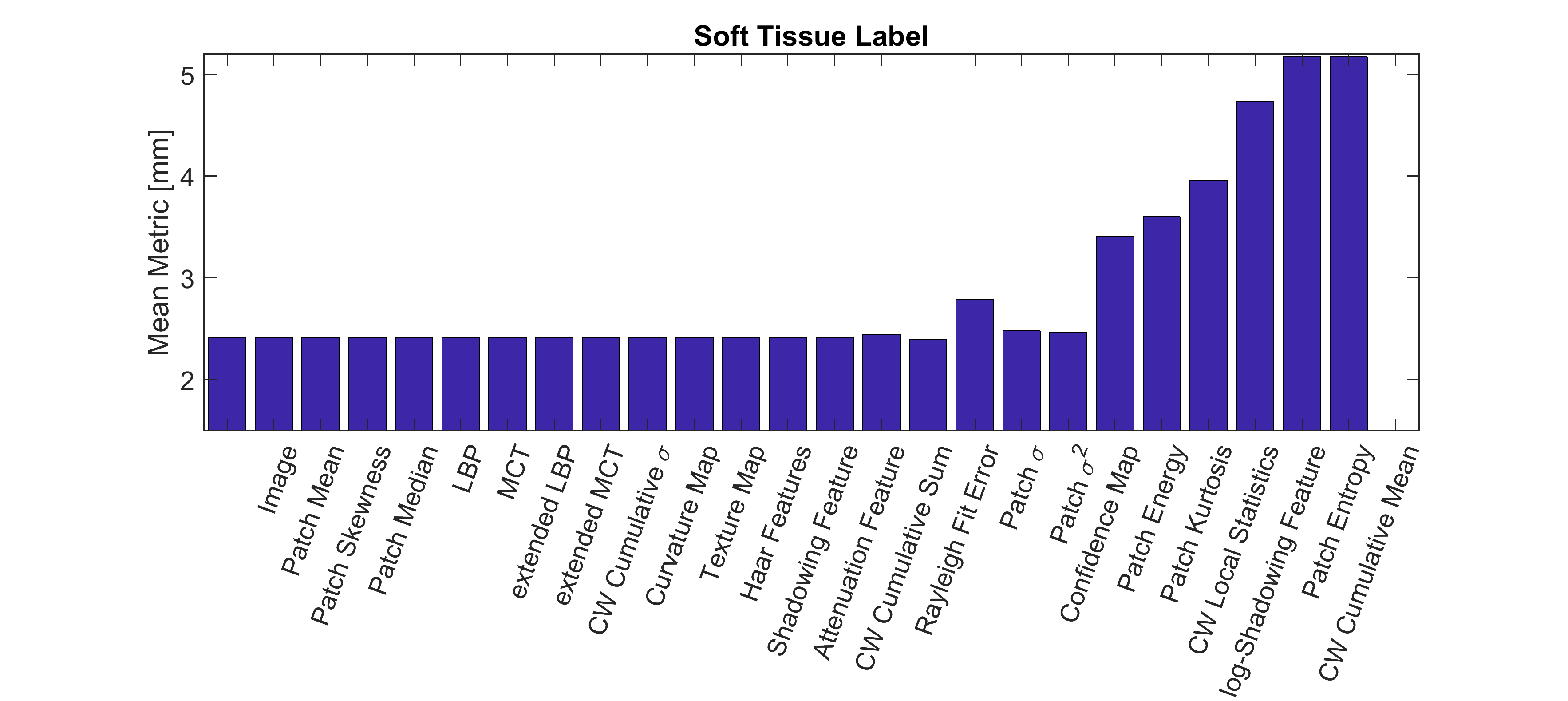}
}
\caption{Change in mean metric (mean of RMSE, oHD, sHD) for $\{\mathrm{diverseUS}\} \setminus S_5$ after sequentially removing the least out-of-bag important feature groups for LogitBoost trained classifiers for (top) shadow and (bottom) soft tissue label. 
Both labels have the same mean metric values by design, but the removal order of feature groups do change for the two classifiers. 
}
\label{fig:perfFeatSelect2}
\end{figure}

\begin{figure}
\centering
\includegraphics[trim = 0mm 0mm 0mm 0mm, clip, width=0.95\linewidth]{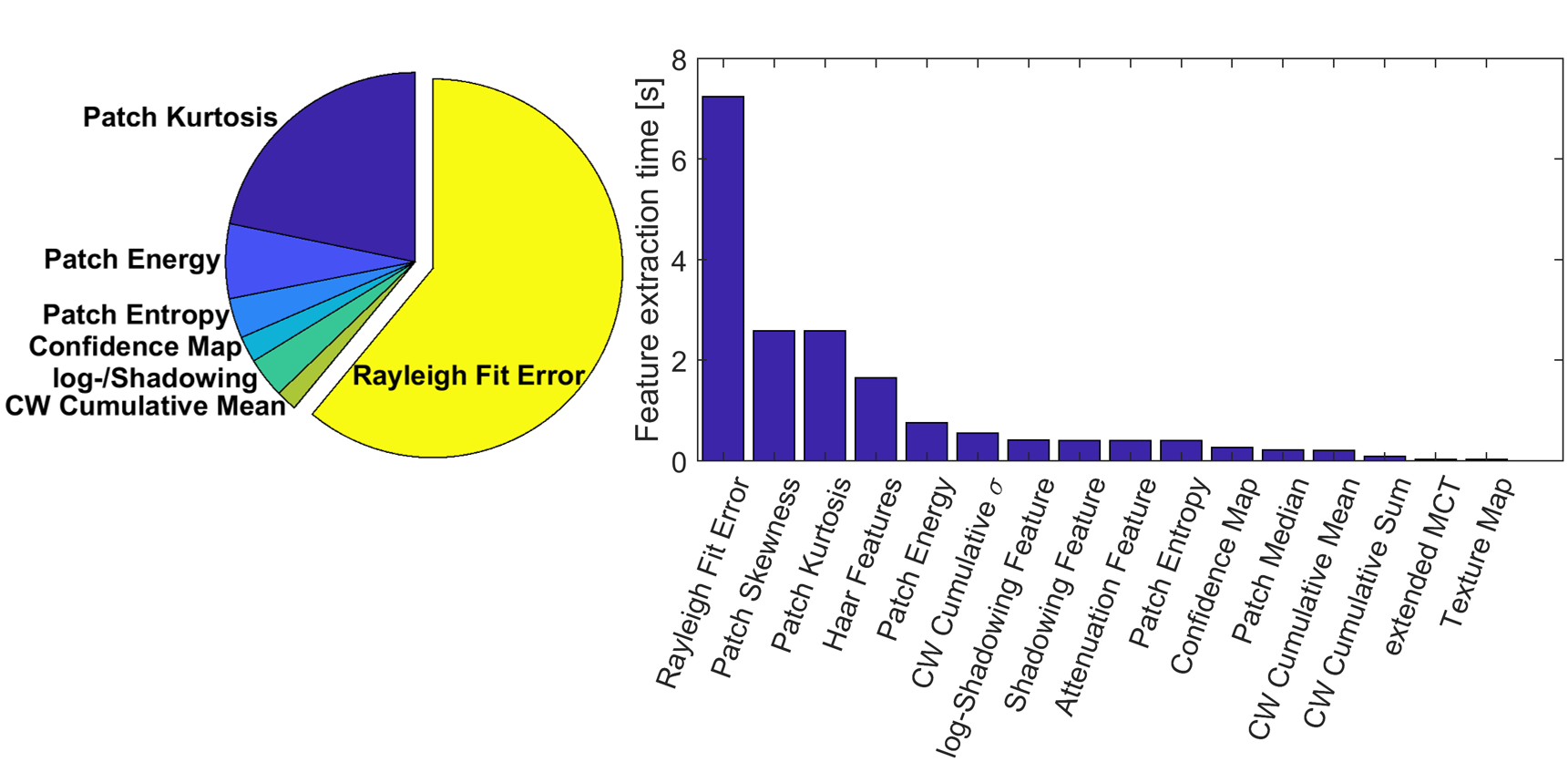}
\caption{
The pie chart (left) depicts the relative computation times of the union of the most important 7 feature groups for both classifiers \textbf{S}~\&~\textbf{T}.
Although it is among the top feature groups, computation time of CW local statistics is not displayed due to being negligibly low compared to the rest. 
Note that log-shadowing and shadowing features are grouped as (log-/Shadowing) as they can be computed together with little overhead.
(Right) Mean computation times of feature groups with $>$30\,ms computation. 
}
\label{fig:featExtractionTime}
\end{figure}

In Fig.~\ref{fig:perfFeatSelect2}, we show the performance drop in the mean metric value (mean$_f$) for the unused samples of diverseUS in subset $S_5$ when greedy feature selection is done for both shadow and soft tissue classifiers. 
Note that removed features vary between the two classifiers. 
The leftmost bar represents mean$_f$ when all feature groups are used, and each consecutive bar indicates mean$_f$ after the removal of the feature group in the axis label underneath that bar.
In addition to the delineation-accuracy based approach for the feature subset selection, one can also check for any potential computation-time based feature (group) elimination for a large speed gain in feature extraction at the expense of minor loss of accuracy.
In Fig.~\ref{fig:featExtractionTime}, we show the feature extraction time for both the most relevant 7 feature groups based on greedy feature selection (cf. Fig.~\ref{fig:perfFeatSelect2}) as well as all feature groups with more than $30$\,ms computation time.
Surprisingly, the union of the most relevant 7 feature groups for the two classifiers (\textbf{S}~\&~\textbf{T}) consist of 8 unique feature groups, which are confidence map, CW cumulative mean, CW local statistics, (log-)shadowing feature, patch entropy, energy and kurtosis, and Rayleigh fit error.

In Fig.~\ref{fig:perfFeatSelect2}, it can be observed that mean$_f$ increases substantially when less than 7 feature groups are left in the feature selection, while it is relatively stationary with more groups. 
Following our intuition earlier, it is not surprising to see that the column-wise cumulative mean feature is among the most descriptive features for both of the classifiers albeit having very little computational footprint. 
Another significant feature for both classifiers is the log-Shadowing feature, which is derived from shadowing feature, used in earlier methods such as CPS~\cite{quader2014confidence}.
Among patch statistics, path entropy is found to be the most relevant for the classifiers. 
This can be attributed to significant difference of the information content between the soft tissue and acoustic shadow.
It can be observed in Fig.~\ref{fig:featExtractionTime}\,(right) that Rayleigh fit error has the largest computational cost while also being of large importance for the classifiers, especially for the shadow label. 
This is followed by patch kurtosis and skewness features having the same feature extraction time, even though kurtosis is found to be substantially more important than skewness. 

One can expect feature importance to vary slightly based on the dataset properties, however, the conducted greedy analysis as well as the average feature computation time can provide intuition for different experiments setups, e.g.,\  applications prioritizing time vs.\ performance.

\end{document}